\documentclass[preprint, 10pt]{elsarticle}

\usepackage{amssymb}

\usepackage{etoolbox}
\PassOptionsToPackage{hyphens}{url}
\usepackage[bookmarks=false]{hyperref} 

\usepackage[T1]{fontenc}
\usepackage{helvet}
\usepackage{float}
\usepackage{microtype}
\usepackage[acronym]{glossaries}

\usepackage[normalem]{ulem}
\usepackage{subcaption,booktabs,siunitx}

\renewcommand{\glossarysection}[2][]{}
\makeglossaries

\journal{Computers and Electronics in Agriculture}
\begin{document}
\fontfamily{phv}
\selectfont
\newacronym{rgb}{RGB}{Red Green Blue}
\newacronym{fads}{FADs}{Foreign Animal Diseases}
\newacronym{rfid}{RFID}{Radio Frequency Identification tags}
\newacronym{kNN}{$k$-NN}{$k$-Nearest Neighbors}
\newacronym{cnn}{CNN}{Convolutional Neural Network}
\newacronym{dnn}{DNN}{Deep Neural Network}
\newacronym{resnet}{ResNet}{Residual Neural Network}
\newacronym{scm}{SCM}{Spatial Context Module}
\newacronym{pointnet}{PointNet}{PointNet}
\newacronym{mlp}{MLP}{Multi-Layered Perceptron}
\newacronym{bcs}{BCS}{Body Condition Score}
\newacronym{gradcam}{Grad-CAM}{Gradient-weighted Class Activation Mapping}
\newacronym{pcsm}{PC-SM}{Point Cloud Saliency Mapping}
\newacronym{tsne}{t-SNE}{t-distributed Stochastic Neighbour Embedding}

\begin{frontmatter}
\title{Universal Bovine Identification via Depth Data and Deep Metric Learning}
\author[]{Asheesh Sharma\texorpdfstring{$^{\text{a,c,d}}$,}, Lucy Randewich\texorpdfstring{$^{\text{d}}$,}, William Andrew\texorpdfstring{$^{\text{a,d}}$,}, Sion Hannuna\texorpdfstring{$^{\text{d}}$,}, Neill Campbell\texorpdfstring{$^{\text{d}}$,}, Siobhan Mullan\texorpdfstring{$^{\text{a}}$,}, Andrew W Dowsey\texorpdfstring{$^{\text{a,b,1}}$}\texorpdfstring{\fnref{cor1}}{}, Melvyn Smith\texorpdfstring{$^{\text{c,1}}$}\texorpdfstring{\fnref{cor1}}{}, Mark Hansen\texorpdfstring{$^{\text{c,2}}$}\texorpdfstring{\fnref{sen}}{}, and Tilo Burghardt\texorpdfstring{$^{\text{d,2}}$}\texorpdfstring{\fnref{sen}}{}}
\fntext[cor1]{Corresponding authors at: University of Bristol, Langford House, Bristol, BS40 5DU, UK (Andrew W Dowsey: andrew.dowsey@bristol.ac.uk). Centre for Machine Vision, University of the West of England, Bristol, Coldharbour Lane, Bristol, BS16 1QY, UK (Melvyn Smith: melvyn.smith@uwe.ac.uk).}
\fntext[sen]{Mark Hansen and Tilo Burghardt are joint senior authors.}
\address{$^{\text{a}}$Bristol Veterinary School, University of Bristol, Langford House, Bristol, BS40 5DU, UK\\
$^{\text{b}}$Department of Population Health Sciences, University of Bristol, Oakfield House, Oakfield Grove, Bristol, BS8 2BN, UK\\
$^{\text{c}}$Centre for Machine Vision, University of the West of England, Bristol, Coldharbour Lane, Bristol, BS16 1QY, UK\\
$^{\text{d}}$School of Computer Science, University of Bristol, Merchant Venturers Building, Woodland Road, Bristol, BS8 1UB, UK}

\begin{abstract}
This paper proposes and evaluates, for the first time, a top-down (dorsal view), depth-only deep learning system for accurately identifying individual cattle and provides associated code, datasets, and training weights for immediate reproducibility. An increase in herd size skews the cow-to-human ratio at the farm and makes the manual monitoring of individuals more challenging. Therefore, real-time cattle identification is essential for the farms and a crucial step towards precision livestock farming. Underpinned by our previous work, this paper introduces a deep-metric learning method for cattle identification using depth data from an off-the-shelf 3D camera. The method relies on \acrfull{cnn} and \acrfull{mlp} backbones that learn well-generalised embedding spaces from the body shape to differentiate individuals -- requiring neither species-specific coat patterns nor close-up muzzle prints for operation. The network embeddings are clustered using a simple algorithm such as \acrfull{kNN} for highly accurate identification, thus eliminating the need to retrain the network for enrolling new individuals. We evaluate two backbone architectures, \acrfull{resnet}, as previously used to identify Holstein Friesians using \acrshort{rgb} images, and \acrshort{pointnet}, which is specialised to operate on 3D point clouds. We also present CowDepth2023, a new dataset containing 21,490 synchronised colour-depth image pairs of 99 cows, to evaluate the backbones. Both \acrshort{resnet} and \acrshort{pointnet} architectures, which consume depth maps and point clouds, respectively, led to high accuracy that is on par with the coat pattern-based backbone. This new universal methodology also addresses the case of all-black and all-white breeds, where the previous coat pattern-based approach fell short. The \acrshort{resnet} colour backbone resulted in $99.97\%$ \acrshort{kNN} identification accuracy, while the PointNet accuracy was $99.36\%$. Our research indicates that these techniques can identify animals using dorsal-view depth maps alone. Regardless of the substantial inter-class variety in the body shape, we show that the models spatially rely on similar body surfaces using \acrfull{gradcam} and \acrfull{pcsm}.
\end{abstract}


\end{frontmatter}

\section{Introduction}
\label{introduction}
In the dairy industry, monitoring each cow is essential for animal wellbeing, environmental sustainability, and farm productivity \cite{McAuliffe2018}. For instance, individual identification is a prerequisite for body condition scoring and yield monitoring that allows farmers to provide tailored care for each cow and boost production \cite{Sumner2018, DeGraves1993, Shahbaz2024}. Continuous identification of individuals is also vital for studying the spread of diseases through contact tracing and social interactions \cite{Disney2001}.

With increasing herd size and the resulting skew in cow-human ratio, tasks such as studying diseases, monitoring yield, and adhering to the best welfare practices are increasingly becoming difficult to conduct. In this study, we examine the effectiveness of utilising body morphology (3D surface characteristics) for individual identification using a commercial depth camera placed at a distance from the animals (non-contact), which requires only one-time installation. This approach aims to transcend the limitations of breed-specific coat patterns, which restricted the scope of our previous research primarily to Holstein-Friesians \cite{Andrew2021}. As per the 2008 report by the \citet{DEFRA2008}, most registered cattle do not have distinct coat patterns. These include $21\%$ Limousin, $9\%$ Charolais, $8\%$ Aberdeen Angus, $8\%$ Simmental, $6\%$ Belgian blue, and $16\%$ other breeds: black, white, or brown. Thus, our primary focus is to establish a universally applicable method for cattle identification by exploring depth as a biometric modality.
\subsection{Motivation}
\label{motivation}
Despite promising advancements in depth-based identification methods for cows \cite{Bezen2020, Okura2019}, several limitations persist: \textbf{1)} The systems often require substantial re-calibration to enrol new animals, \textbf{2)} They depend on combining RGB and depth data, potentially limiting their use to cattle with distinct coat patterns, and \textbf{3)} They may perform unpredictably in unstructured farm environments because of their dependence on accurate estimation of secondary information such as gait.
\begin{figure}[!b]
\centering
\includegraphics[width=1\columnwidth]{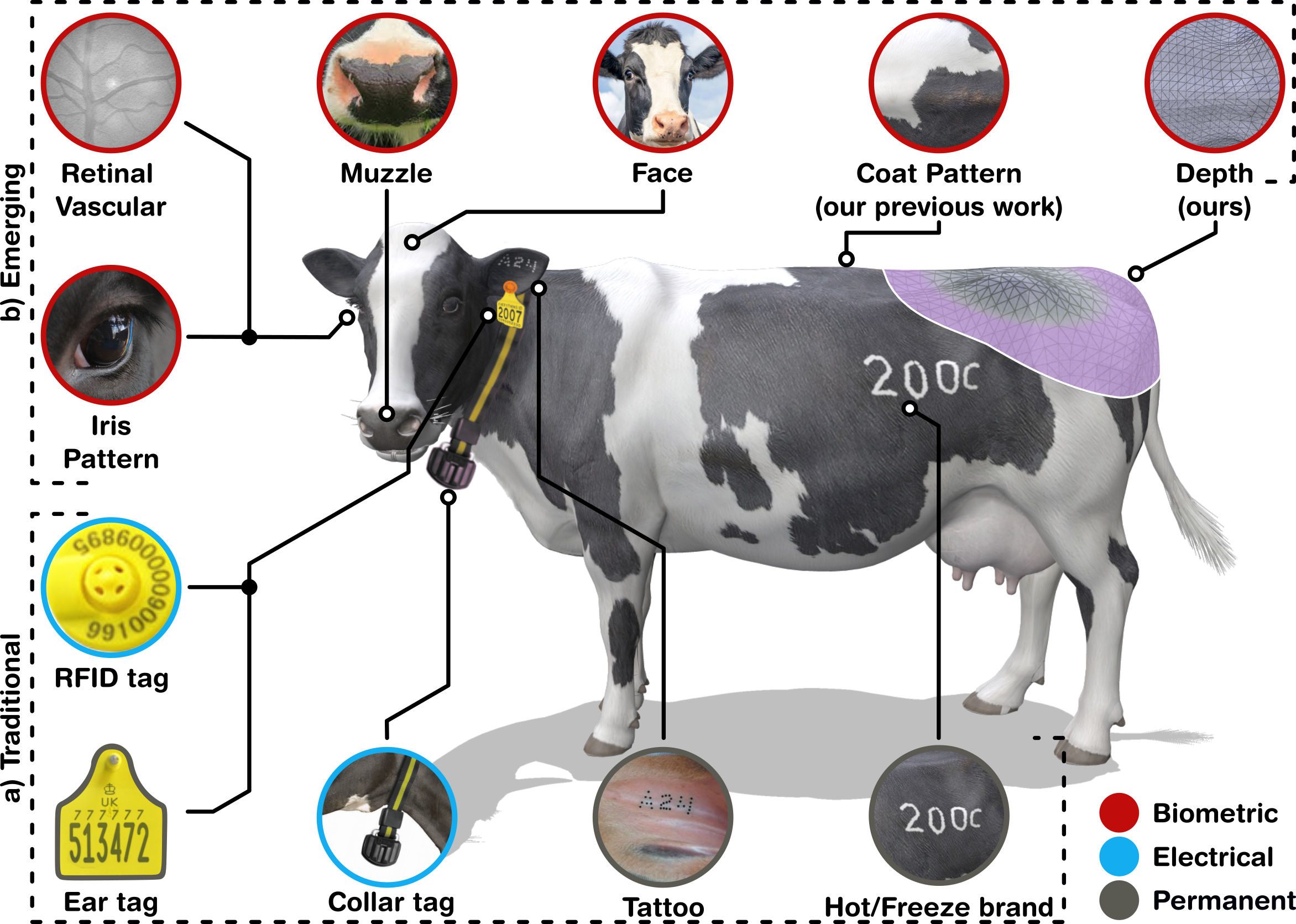}
\caption{{{\label{id_methods}
\textbf{Traditional and emerging identification techniques: Depth is a universally applicable, contact-free biometric for cattle identification.} \textbf{a)} Hot/cold branding, tattooing, and RFID ear/collar tags are traditional methods for identifying cattle. \textbf{b)} Recent research into techniques incorporating biometrics like face, iris, and coat patterns has been made possible by breakthroughs in deep neural networks. Our method employs 3D surface (depth) features for contact-free cattle identification.}}}
\end{figure}
Among permanent identification techniques, \acrfull{rfid} is the current standard which enables wireless tracking of individuals (see Fig. \ref{id_methods}.a) \cite{Stankovski2012}. Despite its widespread use, implementation of \acrshort{rfid} based identification can be expensive and time-consuming, with costs ranging from $\$0.41/$head to $\$5.95/$head depending on the size of the herd and the age of animal \cite{Adam2016}. Furthermore, \acrshort{rfid} based techniques are susceptible to challenges such as data loss, signal distortions, and breakage \cite{Awad2016}.

Besides the cost of procurement and maintenance, traditional systems (see Fig. \ref{id_methods}.a) often have several welfare challenges. For instance, research by \citet{Johnston1996} indicated that just $2.9\%$ of ears with metal tags were not permanently harmed. In a review, \citet{Chapa2020} reported that body sensors and collar tags could detrimentally impact the cow's health. \citet{Lind2018} studied the impact of a tail-mounted calving alert system (Moocall) on the behaviour of pregnant cows as observed through interviews with the farmers. $80\%$ of the farmers observed signs of discomfort in cattle, and most farmers observed physical damage to the tail. Some studies have shown that sensor tags can cause unnoticed abrasive wounds, especially for growing animals, which need adjustment occasionally \cite{Ledgerwood2010, Kokin2014}.

A machine vision approach, coupled with camera systems for identification, can overcome such financial and welfare barriers by providing continuous identification over extended periods without being physically obtrusive. Fig. \ref{id_methods}.b shows emerging machine learning methods to identify cows using bio-metric features such as the face \cite{Li20222}. \citet{Kumar2018} explored the use of muzzle features with great success in terms of accuracy metrics, reaching an identification accuracy of $98.7\%$. Iris and retinal vascular patterns are other examples of such bio-metric features \cite{Lu2014, Allen2008}. However, capturing high-quality images of these features is often difficult. For example, iris images are challenging to capture because of reflections or the occlusion caused by eyelashes. Capturing clear muzzle images requires the animal to stay still and close to the camera system. Despite these challenges, individual identification with these methods has been successful, albeit in research settings under controlled image capture conditions.
\subsection {Open-set identification via deep metric-learning}
\begin{figure}[!b]
\centering
\includegraphics[width=1\columnwidth]{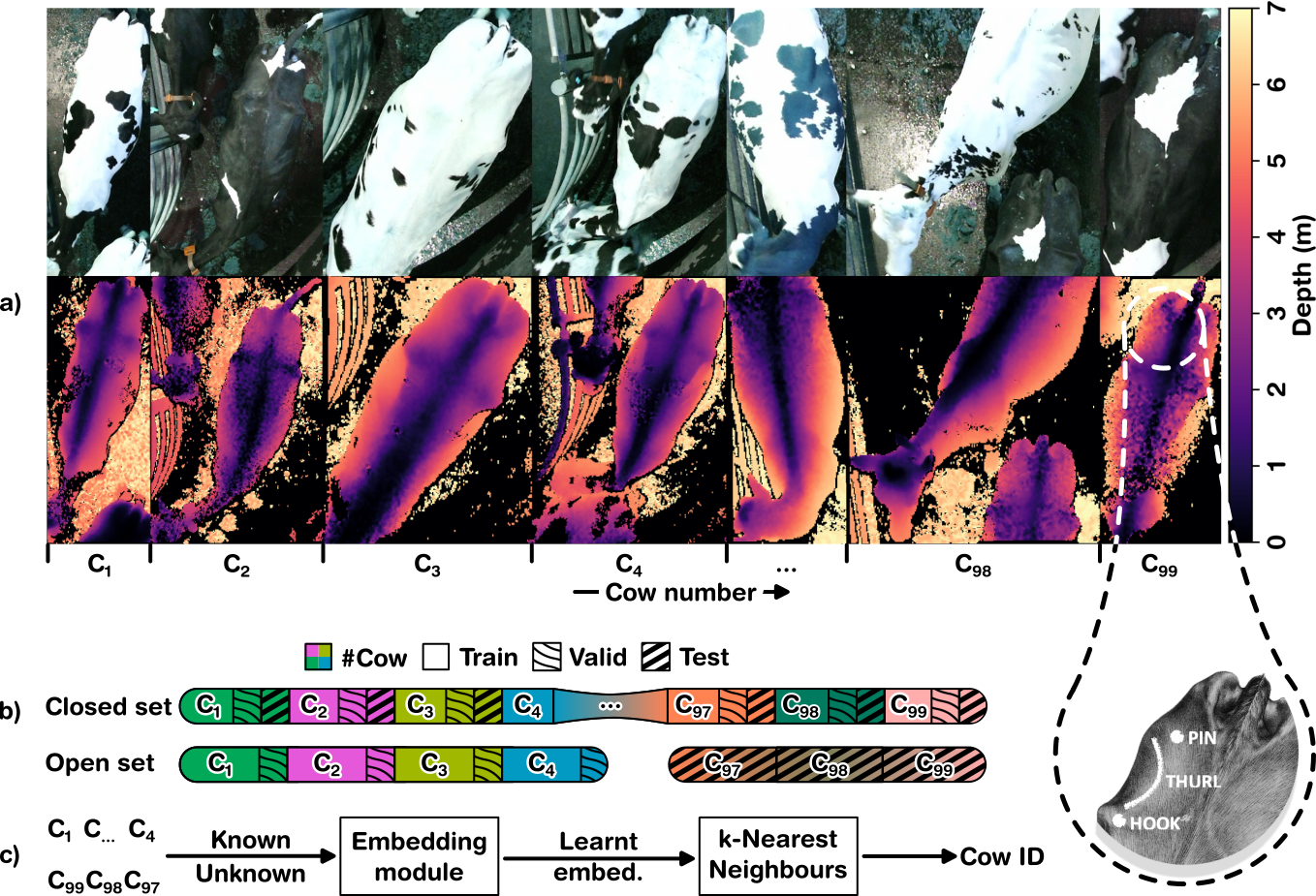}
\caption{{{\label{system_overview}
\textbf{Open-set identification with depth data via deep metric learning.} \textbf{a)} Examples of top-down RGB+Depth images in our dataset, pairs correspond to the same animal, column-wise. The inset shows the hook-pin-thurl region of a cow. \textbf{b)} The open-set validation excludes individuals completely, while the closed-set excludes some portion of the dataset for all individuals. \textbf{c)} A deep learning model generates embeddings clustered using \acrshort{kNN} for identification.
}}}
\end{figure}

We address the challenges mentioned in section \ref{motivation} by employing a top-down (dorsal view) imaging approach, which involves placing a camera at a distance ($\approx 4$ meters) from the animals to enable the identification of multiple individuals simultaneously (see Fig. \ref{system_overview}a). Our previous work, further discussed in section \ref{metriclearning}, has successfully identified Holstein-Friesian cattle through dorsal-view \acrshort{rgb} imagery. The limitation encountered was identifying black cows, which lack distinctive coat patterns \cite{Andrew2021}. Thus, this study investigates whether body morphology, visible in the top-down 3D imagery, such as the spine and the hook-thurl region (Fig. \ref{system_overview}a; inset), are suitable for Holstien-Friesen cattle identification without observing coat patterns. Furthermore, this study opens the door for the individual identification of cattle breeds without coat patterns: 98\% of beef, 49\% of dairy and 17\% dual-purpose cattle in the U.K. (see p. 18, \cite{DEFRA2008}).

To illustrate the broader applicability of our framework, we first describe the process used for assessing the performance of \acrfull{dnn}s utilised in this study. Commonly found in literature, closed-set validation (see Fig. \ref{system_overview}b) involves training and validating \acrshort{dnn}s on distinct subsets of a labelled dataset. Subsequently, the best model is determined based on their performance on a hand-picked test set. As illustrated in Fig. \ref{system_overview}b (top), the training, validation, and test sets must individually have occurrences of all labels in the dataset. Due to this partitioning scheme,  models trained on a closed set have high classification accuracies. However, their widespread use is limited, especially in real-world cases where the number of labels increases or decreases over time, necessitating periodic retraining of the models on an updated dataset. The retraining becomes impractical in the case of farms, which experience regular replacement due to calving and culling. Furthermore, a closed-set identification system would be even more unsuitable for enrolling and recognising wild animals.

As shown in Fig. \ref{system_overview}b (bottom), open-set testing offers a more realistic approach where the dataset is split into train and validation sets using the same partitioning scheme. However, the test set is now disjoint from the training and validation sets, i.e., labels in the test set do not occur in validation/train sets. Therefore, the open-set scheme simulates how the model would perform in a natural setting, where it must be capable of identifying unknown individuals. Further details on the open-set partitioning scheme are presented in section \ref{section:datasetprep}, where we describe the dataset used in this study.

A classification model, trained on a closed set, has a fixed number of outputs for predicting the label for an input. Technically, the algorithms and \acrshort{dnn}s employed to learn the identification labels are often developed in a fully supervised setting (i.e., a fixed set of individuals), which is unsuitable for `in the wild' identification because they require re-calibration or re-training as the number of animals on farm fluctuates. Therefore, as the number of labels in an open set is dynamic, we cannot formulate the training process as a traditional classification task. The other common issue associated with a classification-based approach is confidence flickering of \acrshort{cnn}s when applied to videos because such models lack temporal context \cite{Eilertsen2019}. Even though it is a fair assumption that a frame's content barely changes in the subsequent frames, the single image-based model is prone to abrupt misidentification or flickering.

Our work (see Fig. \ref{system_overview}c) presents a flexible identification system, which transforms the subtle differences in the depth data features into a rich multi-dimensional space, where data points from similar inputs are arranged at a geometrically closer distance. Therefore, any known or unknown individual can be transformed into a unique metric or embedding using this multi-dimensional space, latent in the model. Finally, embeddings generated using this latent space can be clustered using robust algorithms such as \acrfull{kNN} to assign identification labels. Section \ref{training} described the process of metric learning in fuller detail.
\section{Related work}
This section reviews biometric features for bovine identification, such as the face, eyes, muzzle, and coat patterns, and their respective deep-learning approaches. We discuss the limitations of these methods and compare them with our depth-based no-contact approach.
\subsection{Biometrics for bovine identification}
In the context of bovine identification, the face has a broad spectrum of biometric features, ranging from regions such as the eyes or muzzle to the overall face. We discuss these modalities individually and their limitations and finally compare them with our approach.

\textbf{Face for identification.} 
Recent studies by \citet{Xu2022} and \citet{Gong2022} have respectively employed Mobilenet and \acrshort{resnet} deep learning architectures to distinguish individual cows accurately using overall facial features. The Mobilenet model, trained with angular margin loss, demonstrated the ability to discern interclass variations with an accuracy of $91.3\%$ across 2,318 images of 90 individuals \cite{Xu2022}. On the other hand, Gong's \acrshort{resnet}-based model achieved an accuracy of $98\%$ on a dataset of 5,677 images from eight individuals \cite{Gong2022}.

One prevalent issue in face-based identification frameworks is the dependency on single-view images, which limits the network's ability to handle varying poses. Addressing this challenge, \citet{Yao2019} compiled an extensive dataset of 51,151 images representing 200 individuals under diverse poses. Using the PnasNet-5 architecture built on top of \acrshort{resnet}, they attained a $94.1\%$ accuracy. Similarly, \citet{Weng2022} introduced a method that integrates grid-based motion statistics with features from a random sampling algorithm to bolster the robustness of identification across multiple cattle breeds, with varying views and lighting conditions, using a dataset of 8812 images from 82 individuals. While these studies attempted to tackle the problem by diversifying their datasets, some recent studies have focused on improving the model's architecture or training process. \citet{Xu2022-2} implemented a self-attention enhanced \acrshort{cnn} in conjunction with local binary patterns. In contrast, \citet{Yang2023} enhanced a \acrshort{resnet} model by jointly optimising cross-entropy and triplet loss. This approach reduced false positives and maintained robustness to facial pose changes, achieving an accuracy of $98.4\%$ with 2,376 images of 110 individuals.

\textbf{Eye based systems.}
The eye or associated features have been extensively studied for identification. One of the earlier works in this domain was by \citet{Allen2008}, who investigated the unique retinal vascular patterns in the eye's macula for identification. They developed a matching algorithm that achieved $98.3\%$ accuracy over 2,227 images of 869 animals. To enhance the robustness of iris-based methods, reliable segmentation and tracking are necessary. For example, \citet{Zhang2009} employed Sobel edge-based segmentation to analyse the pupil and iris of the eye, while \citet{Zhao2011} developed a similar system to improve traceability in the meat supply chain by tracking the iris and generating a derived bar code. Subsequently, \citet{Lu2014} used previous feature engineering methods and the wavelet transform to encode phase information, which achieved a $98.3\%$ correct recognition rate for 60 iris images of 6 cows. More recently, \citet{Larregui2019} addressed the challenge of capturing clear images under changing lighting conditions. They developed an eye-segmentation algorithm as a preprocessor in the identification pipeline and achieved an Intersection over a Union of 0.89 (dataset: \cite{LarreguiD2019}).

\textbf{Muzzle features.} Small nasal protrusions called beads, which can be round, oval, or irregularly shaped, have a unique biometric signature. \citet{Li2022} exploited such muzzle features to test several deep learning models and achieved identification accuracies of $98.7\%$ on a dataset of 4,923 images of 258 individuals (dataset: \cite{xiong_2022_6324361}).

See \citet{Cihan2023}, \citet{Hao2023} and \citet{Qiao2021} for a detailed review of identification methods involving the biometrics mentioned above.

\textbf{Discussion.} Despite the promising results obtained in the studies mentioned above, there are several limitations in using the face or similar biometric traits. Acquiring high-quality iris images can be challenging due to the small region of interest, issues with reflections or eyelashes, occlusions, and the high likelihood of motion-induced blur. Muzzle images are time-consuming to capture as the system discards many images due to low quality or the movement of animals \cite{Li2022}. Images of the face or its aspects are also challenging to acquire due to the proximity of the animals to the imaging equipment. Aside from logistical challenges, such systems may require physically restraining the animal, which raises welfare concerns.

In contrast, our study takes a non-contact approach of capturing top-down depth images, where the chances of occlusion or loss in quality are less likely. Furthermore, a gap in longitudinal analysis exists, where the reliability of the identification system is tested on datasets collected over extended periods. Our work attempts to fill the research gap by partitioning the dataset for training and testing based on different time scales (see section \ref{section:datasetprep}).
\subsection{Coat patterns and metric learning}
\label{metriclearning}
Top-down imagery of cows is easier to capture because the equipment can be placed unobtrusively and away from the animal. Instead of learning the classification labels for every individual, a set of features can be learned that represents the image's characteristic pattern. The features or latent space can then be clustered into similar-looking images using simple algorithms such as \acrshort{kNN}, thereby eliminating the need for re-training. The approach of learning a latent space is known as metric learning and has demonstrated robustness for identifying unknown Holstein-Friesian cows based on their coat patterns. \citet{Gao2022} developed a novel labelling framework for cattle identification using self-supervised metric learning, producing a test accuracy of $92.4\%$ with just a few minutes of human labelling (dataset: \cite{Neill2021}). In their previous labelling efforts, \citet{Gao2021} used Gaussian Mixture Models and modified RetinaNet to extract and automatically label regions of interest on the publicly available Cows2021 dataset \cite{Neill2021}. \citet{Andrew2021} presented the OpenCows (2020) dataset and used the same methodology to identify Holstein Friesian cattle, producing $93.8\%$ accuracy (dataset: \cite{OpenCows2020}). In their previous study, \citet{Andrew2017} developed the AerialCattle (2017) dataset and used a VGGNet to classify individuals with $97\%$ accuracy (dataset: \cite{AerialCattle2017}). \citet{Andrew2019} developed an autonomous unmanned aerial vehicle system with region scanning and spatiotemporal identification capabilities for Holstein-Friesian cattle in a free farm setting. \citet{Zhao2022} proposed the compact loss function, which improved the identification accuracy by forcing the model to learn tighter clusters and achieved $98\%$ accuracy on the OpenCows (2020) dataset. More recently, \citet{Lu2023} extended the \acrshort{resnet} backbone with a Local Key Area module to improve the identification accuracy to $95.4\%$ and $91.58\%$ on OpenCows (2020) \cite{OpenCows2020} and Cows (2021) \cite{Neill2021} datasets, respectively. 
\subsection{Capturing depth data and related analysis: }
To the best of our knowledge, while depth data has garnered attention for assessing animal health, its full potential for identification purposes has yet to be fully explored. For instance, in our past work \citet{Hansen2018}, we captured depth data from a Kinect sensor to achieve a \acrfull{bcs} repeatability of $\pm$ 0.25. \citet{Alvarez2018} used the SqueezeNet architecture on depth images to achieve $94\%$ classification accuracy of \acrshort{bcs} on a one to five scale with a 0.5 interval. On the same \acrshort{bcs} scale, \citet{Yukun2019} used a CNN architecture to classify individuals with $97\%$ accuracy. Similarly, \cite{Huang2019} used transfer learning with Faster Region-based CNN to achieve $70\%$ classification accuracy for \acrshort{bcs}. \cite{Fischer2015} captured 3D surface geometry from Xtion Pro live motion sensors and developed a regression model that achieves a repeatability of $\pm$ 0.28. \citet{Liu2020} developed an ensemble Gaussian mixture model that can locate various regions of interest within the depth data and achieved $94\%$ classification accuracy on a (1, 5) \acrshort{bcs} scale.

The methodology we used to collect depth data and build our dataset is comparable to the aforementioned approaches. Among other recent data collection methods, \citet{Kadlec2022} developed a fully automated system to capture dorsal-view depth images of cows using contour detection and segmentation algorithms, which is the closest in similarity to our methodology. In section \ref{section:datasetprep}, we describe the method used for data collection in detail.

\section{Methodology}
\label{datasetprep}
This section describes the methodologies for \textbf{1)} Building the COWDepth2023 dataset, \textbf{2)} The \acrshort{dnn} architectures (\acrshort{resnet} and \acrshort{pointnet}) used to generate the embeddings for \acrshort{kNN} based identification method, and \textbf{3)} A technical description of the training setup.
\subsection{Dataset preparation}
\label{section:datasetprep}
\begin{figure}[!t]
\centering
\includegraphics[width=\textwidth]{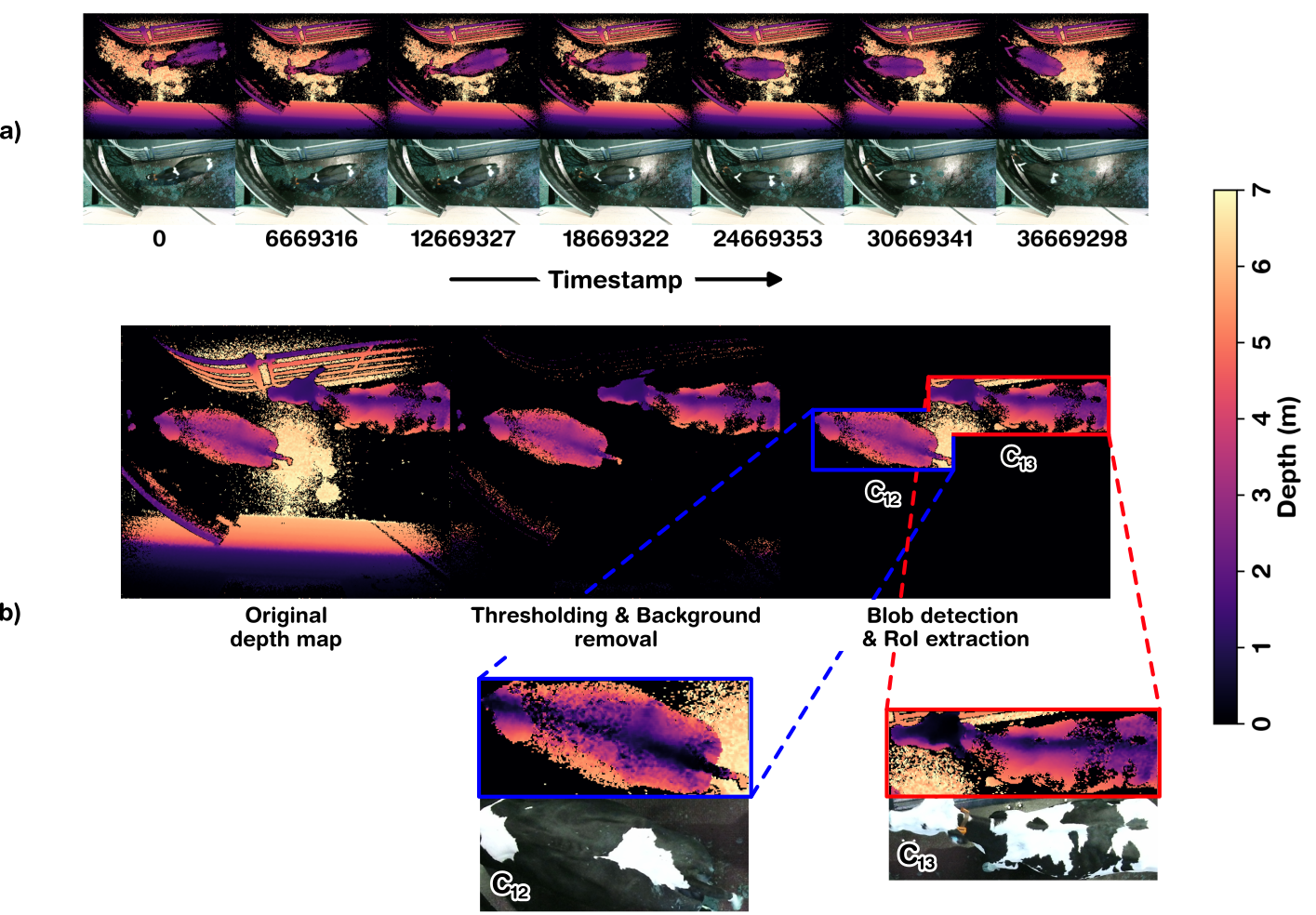}
\caption{\textbf{The camera's point of view and post-processing of depth images.} \textbf{a)} The progression of one cow travelling toward the milking parlour (from left to right) is displayed as a sequence of images and the corresponding time-synchronised depth maps. \textbf{b)} The original depth map is segmented by first thresholding and background subtraction, followed by ``blob'' detection and cropping.}
\label{figure:cow-path}
\end{figure}

We introduce CowDepth2023, a new public\footnote{\href{data.bris.ac.uk}{data.bris.ac.uk}} dataset for studying depth as a feature for identification featuring 21,490 images of 99 Holstein-Friesian cows. We recorded 16-bit depth and 8-bit \acrshort{rgb} streams and frame-level device timestamps with the Kinect V2 camera, which uses infrared light patterns reflected off objects \cite{Han2013}. The device timestamps were later utilised for associating depth and \acrshort{rgb} frames, i.e. time-synchronisation. We used the camera under the default settings, with a resolution of 480i at 30Hz, a horizontal/vertical field of view of $70^o\times 60^o$, at an operating range of 0.5 to 4.5 meters. We collected 14 data sequences, totalling around 50 gigabytes over a duration of 14 days. Fig. \ref{figure:cow-path}a illustrates a time-synchronised example of the captured \acrshort{rgb} and depth sequence, where the camera was placed facing down and aimed towards the ground ($\approx 4$ meters), as the cow walked across the frame.

\textbf{Preprocessing.} We developed our cattle segmentation pipeline based on the \acrshort{rgb} coat pattern identification method introduced by \citet{Andrew2016}. First, the depth images were thresholded to exclude measurements outside the 2 to 3.4 meters range (see Fig. \ref{figure:cow-path}b), which is approximately the height of a Holstein-Friesian cow. Then, structures like gates that were part of the background were removed from the frame. The process involves a background depth image of the scene (with no cows) where any values outside the (1, 3.8) meter range are set to zero to isolate the mid-range depth values corresponding to the cows effectively. Then, any values corresponding to the background are discarded in a target depth image to isolate the cows. After background subtraction, a median filter was applied to smooth each image, filling small holes caused by depth measurement discrepancies. Finally, the largest regions or `blobs' were extracted using SKLearn's label function \cite{scikit-learn} and converted to bounding boxes (see Fig. \ref{figure:cow-path}b, inset). The process resulted in one folder for each depth frame, containing the segmented raw depth/\acrshort{rgb} pairs for individual cows, ready for the labelling process. For all properties of the dataset and the parameters used for preprocessing, refer to Table \ref{table:dataset_props} or the code repository\footnote{\href{https://github.com/Asheeshkrsharma/RGBD-Bovine-Identification}{https://github.com/Asheeshkrsharma/RGBD-Bovine-Identification}}. 

\begin{table}[tb] 
    \small
	\centering 
	\begin{tabular}{p{0.4\textwidth}p{0.46\textwidth}} 
		\toprule 
		\textbf{Parameter/Property}               & \textbf{Description/Value}               \\
        \toprule 
		Image Count                               & 21,490 images, 99 Holstein-Friesian cows \\ 
		Camera Type                               & Kinect V2, using infrared light patterns \\ 
	    Image Types                               & 16-bit depth and 8-bit RGB streams       \\ 
        Camera Resolution                         & 480i, operating at 30Hz                  \\ 
        Total Data Size                           & Approximately 50GB                       \\ 
        Depth Image Threshold                     & (2, 3.4) meters                          \\  
	    Background Image Threshold                & (1, 3.8) meters                          \\ 
		Median Filtering size                     & 4                                        \\    
        Blob area threshold        & (8000, 22000) pixel$^{2}$                         \\    
		\midrule 
		\bottomrule
	\end{tabular}
	\caption{\textbf{Properties of CowDepth2023 dataset.} For further details, refer to the code repository.}
	\label{table:dataset_props}
\end{table}

The process of labelling images containing more than one cow was challenging as the algorithm would often group cows in a single bounding box, resulting in the loss of segmented images. In such cases, we visually compared the coat pattern of the mislabelled cows with the correctly labelled portion of the dataset and reassigned them to a new or existing category.

\begin{figure}[H]
\centering
\includegraphics[width=1\textwidth]{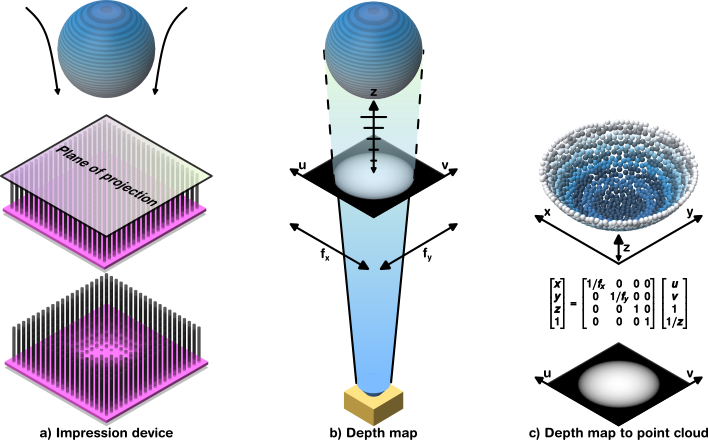}
\caption{\textbf{Analogy for interpreting depth} \textbf{a)} Pushing an object against the pin impression device creates depressions at the contact plane; depth maps represent a similar property. \textbf{b)} The depth camera measures the distance between itself and the object for each pixel. \textbf{c)} Point clouds are generated using the depth map and the camera's intrinsic properties that associate individual points to the measured depth.}
\label{figure:toyimpression}
\end{figure}

\textbf{Depth maps to point clouds.} The difference between a point cloud and a depth map must be highlighted before discussing how these representations are converted back and forth. A depth map resembles a 2D rendering of an object's surface, reminiscent of the impression left when an object is pressed against a pin impression device (see Fig. \ref{figure:toyimpression}a). The displaced pins are analogous to value coordinates in the depth map. Depending on the density of these pins, the resulting impression captures varying levels of detail from the object's surface.

However, unlike a pin-impression device, which uniformly records the object's features, a camera subtly distorts measurements along its pixel coordinates, represented as (u, v), influenced by its focal length and lens distortion characteristics (see Fig. \ref{figure:toyimpression}b, $f_x, f_y$). Another distinction is that while the device develops an impression precisely at the plane of contact with the object, a camera's plane is farther away. As a result, the camera's intrinsic properties are required to precisely map between pixel locations (u, v) and lengths (x, y). A point cloud is the representation we get when a depth map is remapped from pixel coordinates (u, v, distance) to natural unit coordinates (x, y, distance). The mathematical foundation for the remapping is out of the scope of this study; refer to \citet{Hossain2023} for further details. Finally, the above process resulted in raw point clouds with varying numbers of points, often too large to be used directly for training. Therefore, using the farthest point sampling algorithm \cite{Eldar1994}, we uniformly resampled the point clouds to a fixed number of points ($2^{14}$) as recommended by \citet{Qi2016}.

\begin{figure}[H]
\centering
\includegraphics[width=1\textwidth]{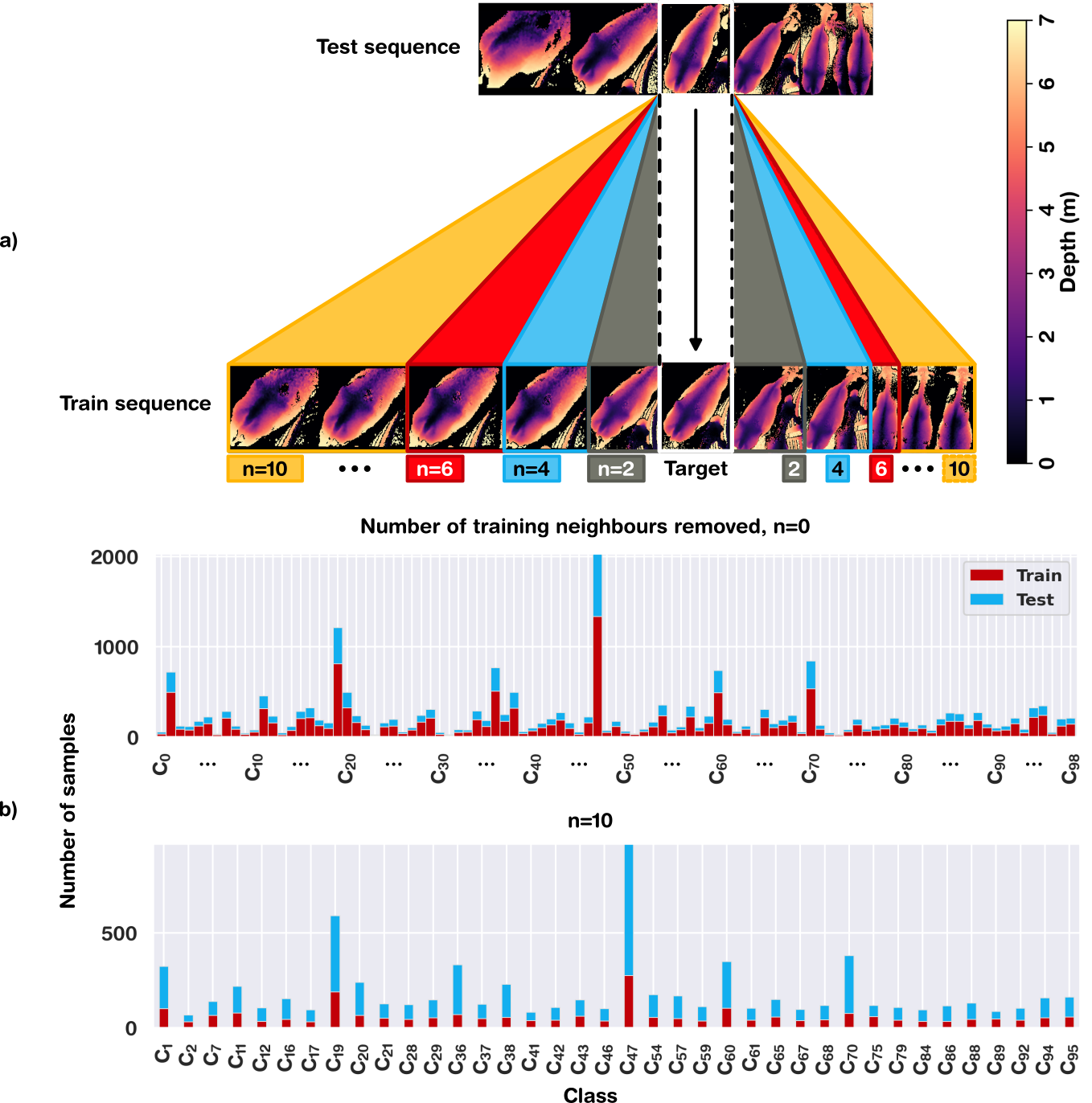}
\caption{\textbf{Preparing the dataset for open-set analysis.} First, we randomly split the dataset into train and test sets in a 7:3 ratio. \textbf{a)} Then we discard n neighbouring images from the train set for every test image. \textbf{b)} As the value of n is increased from 2 to 10, more images are discarded from the train set, resulting in some cows with no images left. These cows in the test set are considered `unknowns', i.e. part of the open-set, enrolled with \acrshort{kNN}, post-training.}
\label{figure:dataset_split}
\end{figure}

\textbf{Splitting the dataset for open-set analysis.} Due to the sequential nature of the dataset, it was necessary to explore the effect of marginal differences between temporally neighbouring examples in the training set on the test accuracy. So-called ``leave-sequence-out'' training involves discarding a specified number of temporally neighbouring images (n in Fig. \ref{figure:dataset_split}a) on either side of every test (target) image from the training set. By removing neighbouring samples, we aim to prevent the likelihood of a test example having a temporally similar image in the training set that the model has already learned. 

A consequence of eliminating temporal bias is that as the number of neighbouring images (n) to be removed increases, the overall number of images left in the training set decreases, which, in some cases, naturally reduces to zero for some cows (compare Fig. \ref{figure:dataset_split}b; n=0 and n=10). With no examples in the training set, such cows are considered unknowns that become part of the open set, enrolled only after training the model, with \acrshort{kNN}. Thus, the overall training process simulates part of the real-world scenario when a farm experiences growth in the number of animals. More details on the choices of n, number of unknowns and images in the train and test set are listed in Fig. \ref{figure:nvsaccuracy} of section \ref{section:results}.
\subsection{Model Architectures}
Traditionally, the 3D depth information requires some transformation into sparse, meaningful features before being used in various 3D recognition tasks. The most prevalent reason for the need for this preprocessing step is the computational cost of directly operating on the raw 3D point data. Such features encompass local and global descriptors such as Fast Point Feature Histograms, Viewpoint Feature Histograms, Clustered Viewpoint Feature Histograms, Radius-based Surface Descriptors, and Rotation-Invariant Feature Transform, which are often used in combinations as inputs to 3D scene understanding algorithms for better accuracy \cite{Rusu2009, Rusu2010, Marton2010, Wu2017}. However, determining the ideal recipe for combining such features is a non-trivial process, which depends on the domain and the task at hand. Furthermore, such features perform poorly in cases of noise and occlusion (see \citet{Han2023} for a full review).

\begin{figure}[!b]
\centering
\includegraphics[width=1\textwidth]{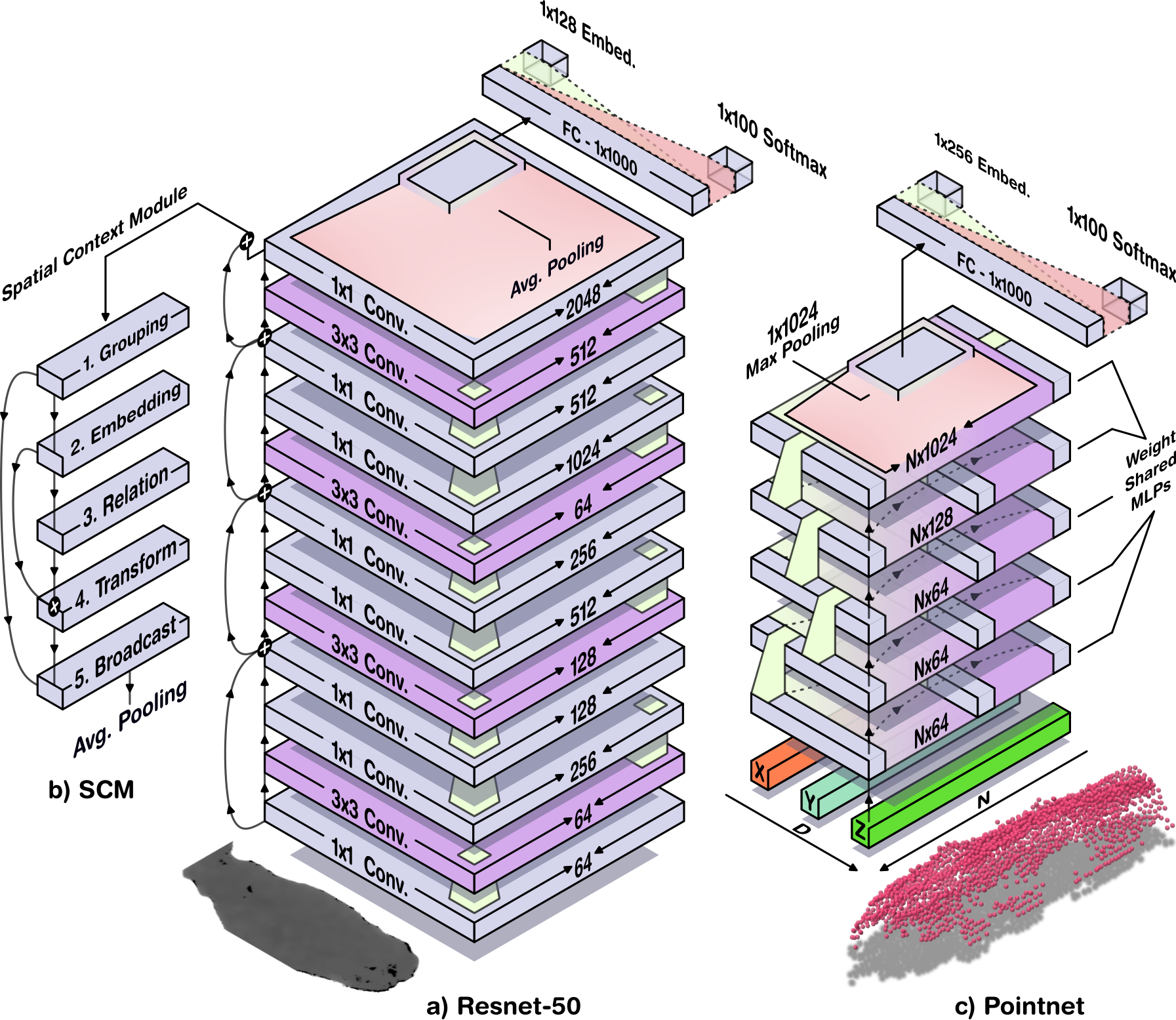}
\caption{\textbf{Network architectures} \textbf{a)} The \acrshort{resnet}-50 and \acrshort{pointnet} architectures consume depth maps and point clouds as inputs, respectively. The  \acrshort{resnet} network consists of 4 convolution blocks; the double arrows between the blocks represent the skip-connections \citet{He2016}. \textbf{b)} The output from the last convolution block is fed to the \acrfull{scm}, followed by the average pooling layer, that generates the final embedding. \textbf{c)} The \acrshort{pointnet} architecture directly consumes a point cloud. The network consists of \acrfull{mlp}s, producing NxD outputs consecutively (N is the number of points, D is the number of channels). The max-pooling layer aggregates the output from the last \acrshort{mlp} layer into a global feature vector.}
\label{figure:resnet}
\end{figure}

\textbf{The \acrshort{resnet} architecture.} Canonical representations such as voxels, meshes, point clouds, and depth maps have recently gained popularity. These representations closely mirror the raw depth data and are employed to circumvent the complexities of feature engineering. In our research, we have focused on depth maps as they directly stem from the camera and align with the established popularity of \acrlong{cnn}s—particularly \acrshort{resnet}s, for which depth maps can act as regular image inputs. The disadvantage of traditional \acrshort{cnn} architectures is their inability to learn more abstract/complex features because of the difficulty in propagating gradients through the network as the number of hidden layers increases. The \acrshort{resnet} architecture alleviates the vanishing gradient problem by employing residual skip connections (see Fig. \ref{figure:resnet}a) between the layers to carry some information from the previous layer as-is to the next, making the gradient descent process more effective at propagating the loss.

In our study, we used the \acrshort{resnet}-50 variant (Fig. \ref{figure:resnet}), using pre-trained ImageNet weights, as proposed by Andrew et al. (2021) \cite{Andrew2021}. As shown in Fig. \ref{figure:resnet}a), \acrshort{resnet}-50 has four convolutional blocks arranged as bottlenecks, followed by a final fully connected (or dense) layer with a (1x2048) output. Each of the four convolutional blocks varies in size, as shown in \ref{figure:resnet}a. In total, there are 49 convolutional layers. The network feeds the dense (1x2048) output from the \acrshort{resnet}-50 to a \acrfull{scm}, followed by the final layers for generating a (1x128) embedding. An additional (1x99) output generates classification labels for the 99 cows in the CowDepth2023 dataset. The classification layer only facilitates the training process for computing the loss functions and does not conflict with the premise of metric learning. In section \ref{training} (training setup), we describe the loss function in conjunction with the purpose of the classification layer.

The \acrfull{scm} allows the network to focus on the most relevant regions of an image by weighing each region's importance for the identification process. We built the \acrshort{scm} into the network before the fully connected layers to boost the model's robustness by focusing on the most critical aspects of each image. As shown in Fig. \ref{figure:resnet}b), the \acrshort{scm} computes feature weights by learning their spatial importance within the context of an image \cite{Yang2019}. The features from \acrshort{resnet}'s last layer (1x2048) are grouped into a spatial context vector, which is forwarded to the average pooling layer. In this case, the \acrshort{scm} can help the model cope with the learning difficulties faced due to occlusion in camera footage. Moreover, it could learn to ignore some apparent anomalies in the images due to imperfections in the camera capture, as discussed in section \ref{section:datasetprep}.

\textbf{The \acrshort{pointnet} architecture.} Due to the inherent 2D nature of depth maps, there is a risk that \acrshort{resnet}s, driven by pattern recognition, may inadvertently learn orientation-related cues. For instance, cues such as the fixed camera orientation can influence the network's generalisation capacity to other novel views. Therefore, we studied point clouds that are constructed by re-projecting the depth data from the camera plane (u, v, distance) to physical units (x, y, distance) using the depth camera's intrinsic parameters (see section \ref{section:datasetprep} for more details). As a result, a trained deep-learning model can consume a point cloud of any unknown orientation.

Our study employs the \acrshort{pointnet} architecture to learn discriminative embeddings from point clouds. In the depth map representation, the semantic meaning of pixels does not generally change with their ordering. In contrast, 3D points carry little information without the context of their neighbourhood. To handle the unique challenges that point clouds pose, \citet{Qi2016} proposed the \acrshort{pointnet} architecture that encodes differently oriented and unordered point clouds to a standard reference by essentially learning a conditional transformation matrix using \acrfull{mlp} networks. Since the goal of the model is to learn the similarity between two point clouds irrespective of the order in which 3D points occur, the output from the last \acrshort{mlp} must be weighted based on relevance (see Fig. \ref{figure:resnet}c). The \acrshort{pointnet} architecture achieves the property of order invariance by using the max operator. This mathematically symmetric function gives the same output regardless of the order of the input arguments. In Fig. \ref{figure:resnet}c, the \acrshort{pointnet} contains four layers of \acrshort{mlp}s with a varying number of output channels. Each \acrshort{mlp} considers every point in the point cloud, producing an Nx3 output, where N is the number of points. Finally, after the fourth layer, the max operator generates a global feature vector (1x1024). To generate the 256-dimensional embeddings, the max pooling output is fed through the additional fully connected layers similar to the \acrshort{resnet} architecture, as discussed before.
\subsection{Training setup}
\begin{figure}[H]
\centering
\includegraphics[width=1\textwidth]{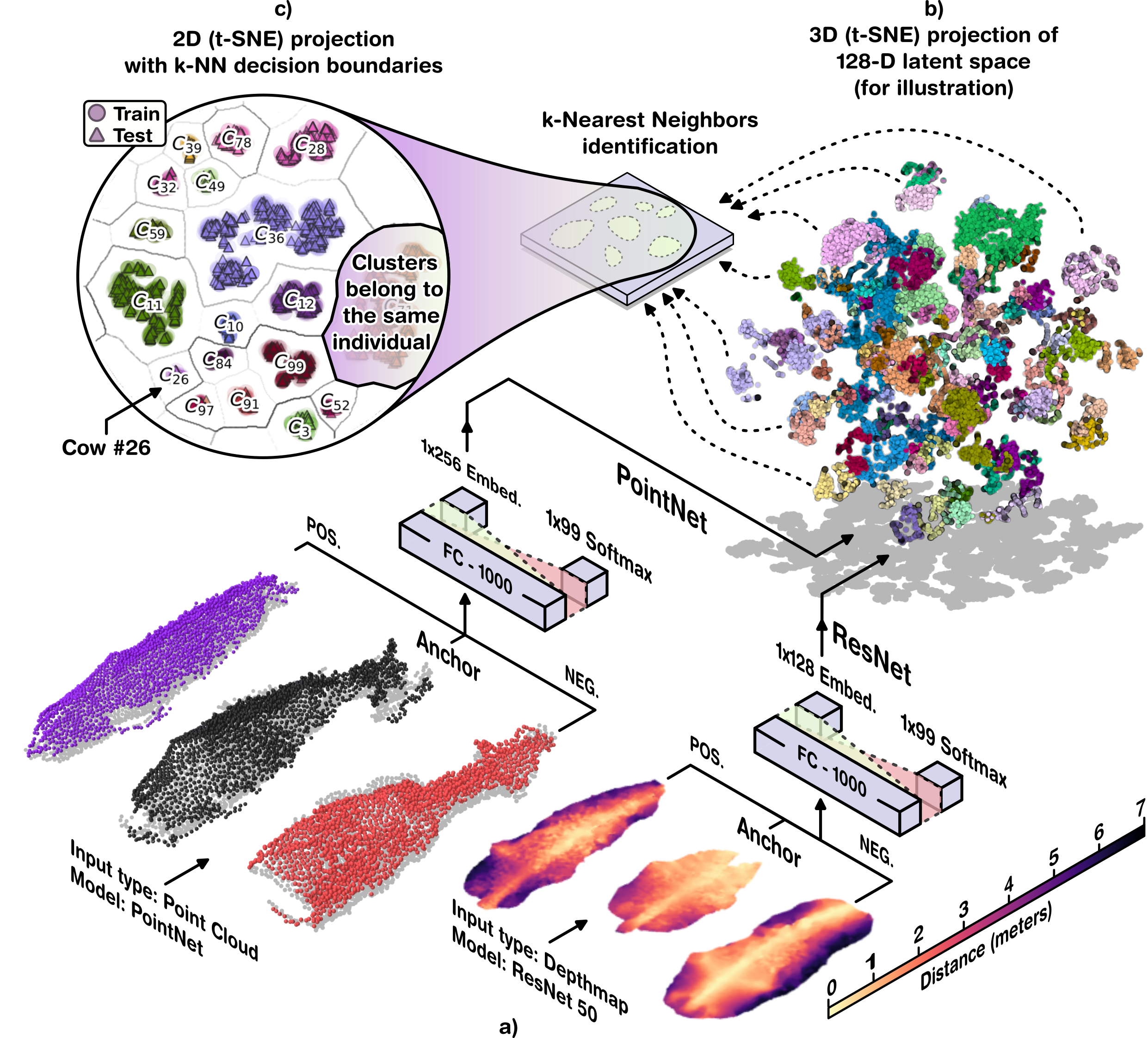}
\caption{\textbf{The metric learning process} \textbf{a)} Depending on the \acrshort{pointnet} or \acrshort{resnet} architecture, a triplet of point clouds or depth maps are used as inputs, respectively. \textbf{b)} This process generates a high dimensional latent space, which is used to train the models, with the central premise being the generation of ID-associated clusters. \textbf{c)} After the training, \acrshort{kNN} is used as a classifier to extract the labels. The latent space cluster visualisations were generated using the \acrshort{tsne} algorithm (perplexity 30) by reducing the high dimensional latent space to 3D and 2D for sub-figures \textbf{b)} and \textbf{c)}, respectively. Black lines in \textbf{c)} represent the \acrshort{kNN} decision boundaries. Note: Some lines appear faded or grey because fewer clusters share the decision boundaries.}
\label{figure:trainingSetup}
\end{figure}

\textbf{Deep metric learning.}
In their study, \citet{Andrew2021} trained the \acrshort{resnet}-50 model (see Fig. \ref{figure:resnet}) to learn a latent space that clusters images from the same individuals with good separation from others. Therefore, the task of learning a discriminative latent space entails penalising the models (\acrshort{resnet} or \acrshort{pointnet}) to maximise the distance between the embeddings of two different animals while minimising the distance between the example depth maps or point clouds of the same individual. As shown in Fig.\ref{figure:trainingSetup}a), both models are presented with three inputs, namely the anchor, positive and negative. After a forward pass, the models generate three 128-dimensional vectors (or embeddings), each corresponding to the three inputs. In the case of well-informed latent space, the anchor embedding should geometrically lie very close to the positive embedding since they belong to the same individual. In contrast, the negative embedding would lie farther apart, thus forming tighter clusters, as shown in Fig. \ref{figure:trainingSetup}b).

\label{training}

\textbf{Loss function.} In deep metric learning, the penalty or incentive is a loss function, which must be selected carefully to promote the models to exhibit the abovementioned properties in their latent space.

First proposed for training siamese networks for predicting the similarity between embeddings ($x_1$, $x_2$) of two images, the contrastive loss function can be formulated as follows:
\[ \mathbb{L}_{Contrastive}=\frac{1-Y}{2}d(x_1, x_2) + \frac{Y}{2}max(0, \alpha -d(x_1, x_2))\]
Where $d(x_1, x_2)$ is the Euclidean distance, $Y$ denotes the image similarity (if $x_1\sim x_2, Y\rightarrow 0$) or dissimilarity ($Y=1$) \cite{Hadsell2006}. The problem with contrastive loss is that it cannot minimise the distance between similar embeddings while simultaneously maximising the distance between dissimilar embeddings. \citet{Schroff2015} overcame this limitation with triplet loss $\mathbb{L}_{TL}$ which uses three embeddings $x_p$, $x_a$, and $x_n$ for positive, anchor, and negative, respectively.

\[ \mathbb{L}_{TL} = max(0, d(x_a, x_p) - d(x_a, x_n) + \alpha)\]
The margin parameter $\alpha$ controls the minimum/maximum distance between image pairs. The triplet loss allows the model to minimise the distance between $x_p$ and $x_a$ while maximising the distance between $x_a$ and $x_n$. The issue with triplet loss is the maximum term, which makes the loss value zero when the distance term becomes negative (i.e. $d(x_a, x_n) > d(x_a, x_p) + \alpha$). The model will not learn to minimise $d(x_a, x_p)$ in such cases. \citet{Masullo2019} proposed the reciprocal triplet loss, which also eliminates the need for $\alpha$:
\[ \mathbb{L}_{RTL} = d(x_a, x_p) + \frac{1}{d(x_a, x_n)} \]
Research has shown that including $Softmax$ can boost the model's accuracy because of the term's supervised nature \cite{Wang2017}. As proposed by \citet{Andrew2021}, the reciprocal triplet loss can be combined with $Softmax$ as follows:
\[ \mathbb{L}_{SoftMax + RTL} = \mathbb{L}_{SoftMax} + \lambda \cdot \mathbb{L}_{RTL}  \cdots{1}\]
where
\[\mathbb{L}_{SoftMax} = - log(\frac{e^{x_{class}}}{\sum_{i}e^{x_i}})\]
Moreover, $\lambda$ is a weighting hyperparameter, for the experiments, we used $\lambda=0.01$ as suggested by \citet{Andrew2021}.

In this study, we used the $\mathbb{L}_{SoftMax + RTL}$ loss function (Eq. 1) to train the \acrshort{pointnet} model and used the following function (Eq. 2) with a similar combination of $Softmax$ and the original triplet loss function in the case of \acrshort{resnet}-50.
\[\mathbb{L}_{SoftMax + TL} = \mathbb{L}_{SoftMax} + \lambda \cdot \mathbb{L}_{TL} \cdots{2}\]

Finally, after the models converge, we use the \acrshort{kNN} algorithm to fit the 128-dimensional embeddings from the training split and produce the identification labels for the test split. We discard the $SoftMax$ output from the models after the training process, and it is not used for identification later on. Therefore, all the results in this study report the classification accuracy for the \acrshort{kNN} algorithm and \textbf{not} for the models themselves.

\textbf{Data augmentation.}
Data augmentation is a widely adopted technique in image recognition systems for diversifying the training data by leveraging image transformations such as rotations and zoom \cite{Shorten2019}. By diversifying the training data, the model's capacity to learn robust features that generalise well to unseen testing data also increases. We used the Augmentor package to produce 60 new images for each cow from random rotation, translation, and cropping of the original images \cite{Bloice2022}. To maintain the aspect ratio of the images, they were padded on all sides in the case of rotation angles divisible by $ 90^o$. The Augmentor package removes these borders by cropping the images to retain the most significant region. This behaviour made the package suitable since creating new artefacts in augmented images would be undesirable. After generating 60 augmented images for each cow using this process, we produced 30 additional images by introducing varying degrees of Gaussian noise with a fixed mean of $\mu=0$ and variance $\sigma^2$ ranging from 10 to 80 \cite{Buslaev2020}. For the point clouds, we used the same augmentation procedures laid out by \citet{Qi2016}; Gaussian noise $(\mu=0, \sigma=0.02)$, along with random rotation along the z-direction.

\textbf{Equipment and Hyper-parameters.}
Given the large amount of data involved in training paired with model complexity, it was imperative to perform experiments on a GPU cluster with CUDA capabilities. All \acrshort{resnet}-50 experiments were executed on the Nvidia Pascal P100 GPUs on the University of Bristol's supercomputer, Blue Crystal (Phase 4) \cite{bc4}. The \acrshort{pointnet} experiments were conducted on a local machine with an Intel Xeon E5-1650 CPU and NVIDIA GeForce RTX 2070 GPU. The hyper-parameters used in this study follow the same values as justified in our previous work \citet{Andrew2021} for \acrshort{resnet}. Similarly, the parameters for \acrshort{pointnet} follow the work by \citet{Qi2016}. The models were set up to evaluate the validation set every other epoch, saving model weights if the previous best \acrshort{kNN} accuracy was surpassed. See Table \ref{table:training_params} for different hyper-parameter choices and their justification.

\begin{table}[htbp] 
    \small
	\centering 
	\begin{tabular}{p{0.9in}p{0.75in}p{0.75in}p{1.7in}} 
		\toprule 
		\textbf{Parameter}               & \textbf{\acrshort{resnet}}           & \textbf{\acrshort{pointnet}}          & \textbf{{Remarks}}                                        \\
        \toprule 
		Optimiser               & SGD                         & Adam                         &\citet{Andrew2021}, \citet{Qi2016}           \\ 
		Learning Rate           & 1e-3                        & 1e-3                         & \citet{Andrew2021}                          \\ 
  Batch Size              & 16                          & 24                           & \citet{Qi2016} \\ 
  Iterations              & 50                          & 50                           & From \citet{Andrew2021}                          \\ 
		Momentum                & 0.9                         & N/A                          & -                                                \\ 
		Mining strategy & Batch hard                  & Batch hard                   & From \citet{Andrew2021}                          \\  
  Weight Decay            & 1e-4                        & 1e-4                         & From \citet{Andrew2021}                          \\ 
		Loss function           & $\mathbb{L}_{S + TL}$ & $\mathbb{L}_{S + RTL}$ & -                                                \\    
		Lambda                  & N/A                         & 0.01                         & From \citet{Andrew2021}                          \\    
		Pre-training      & ImageNet                    & Modelnet40                   & -                                                \\
		\midrule 
		\bottomrule
	\end{tabular}
	\caption{\textbf{Standard hyper parameters for \acrshort{resnet} and \acrshort{pointnet} following \citet{Andrew2021} or \citet{Qi2016}.} All results quoted from experiments in this study were run with these settings unless otherwise stated.}
	\label{table:training_params}
\end{table}

\section{Results}
\label{section:results}
\textbf{Closed-set evaluation.} Firstly, we compared the performance of the \acrshort{resnet} and \acrshort{pointnet} architecture on respective input types without temporal bias removal. The CowDepth2023 dataset with 21,490 depth maps/point clouds of 99 Holstein-Friesian cows was split randomly five times with a ratio of 7:3 such that all cows occur in both train and test sets. The reasons for this preliminary experiment are four-fold: \textbf{1)} To establish a baseline for the two models keeping the temporal bias intact as described in section \ref{section:datasetprep}, \textbf{2)} Compare the performance of \acrshort{resnet}, without the \acrshort{scm} by feed forwarding the output from the average pooling layer directly to the embedding layer as shown in Fig. \ref{figure:resnet}, \textbf{3)} To evaluate the robustness of \acrshort{pointnet} to depth discrepancy, i.e. missing data in depth map, by controlling the number of input points, and \textbf{4)} To find the optimal number of 3D points for the best performing \acrshort{pointnet} variant.

\begin{table}[ht]
	\centering
	\begin{tabular}{lcr}
		\hline
		Model                  & Mean and range        & Diff. from baseline \\
		\hline
    	\addlinespace 
		ResNet-50 (Colour)     & $99.88\%, (-0.13, +0.08)$  & -0.04\%	\\
		ResNet-50-SCM (Colour) & $\textbf{99.91\%}, (-0.13, +0.06)$  & 0.00\%	\\
		\addlinespace 
        ResNet-50 (Depth)      & $\textbf{99.83\%}, (-0.17, +0.07)$  & -0.08\%   \\
		ResNet-50-SCM (Depth)  & $99.82\%, (-0.19, +0.09)$  & -0.09\%   \\
        \addlinespace 
        PointNet (2048 points) & $\textbf{99.09\%}, (-0.70, +0.19)$  & -0.82\%   \\
        PointNet (64 points)   & $87.58\%, (-1.64, +0.62)$  & -12.33\%  \\
        \addlinespace 
        \midrule 
        \bottomrule
    \end{tabular}
	\caption{\textbf{Model performance.} We compare the mean and the range of \acrshort{kNN} test accuracies for various models. We divided the dataset into train-test sets with a ratio of 7:3 and repeated the experiment 5 times. For all the models arranged in rows, the third column reports the difference in accuracy from the best-performing baseline model -- \acrshort{resnet}-50-SCM (colour).}
	\label{tab:closedset}
\end{table}

According to Table \ref{tab:closedset}, the baseline \acrshort{resnet}-50 model achieves a \acrshort{kNN} mean accuracy of $99.88\%$ with \acrshort{rgb} images of coat patterns. In comparison, our \acrshort{resnet}-50 model with depth map as input performs similarly well on the test data, with only $0.05\%$ less accuracy. Furthermore, including the \acrshort{scm} layer does not yield any significant improvement; in the case of \acrshort{rgb} images, the accuracy slightly increases by $0.03\%$, and decreases by $0.01\%$ in the case of depth maps. The performance of \acrshort{pointnet} model is sensitive to the number of input 3D points. Overall, both \acrshort{resnet} and \acrshort{pointnet} models can generalise well on depth maps and point clouds, respectively, and exhibit performance that is comparable to models that use \acrshort{rgb} images of coat patterns.

\begin{figure}[!t]
\centering
\includegraphics[width=1\textwidth]{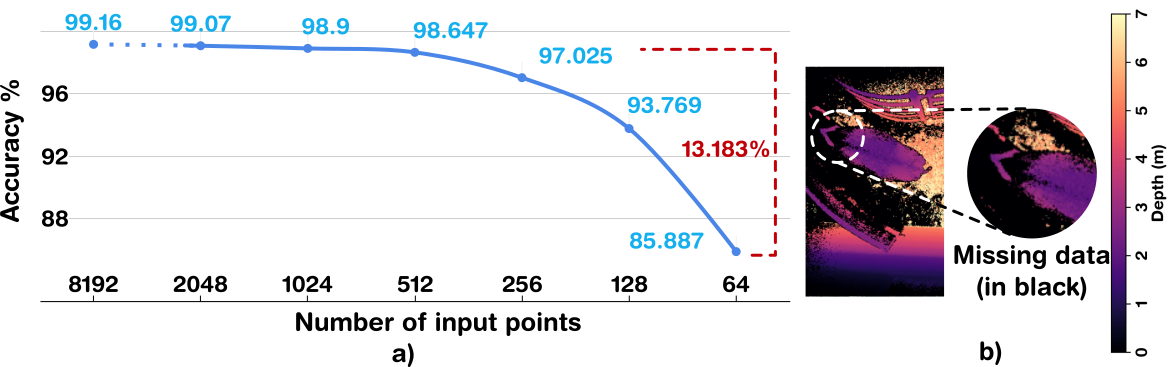}
\caption{\textbf{Robustness of \acrshort{pointnet} to missing data.} Depth cameras produce areas of missing measurements as illustrated in the depth map of \textbf{b)}. In \textbf{a)}, we quantify that \acrshort{pointnet} is robust to a significant amount of missing depth data. Representing individuals by 64 3D data points still leads to $\approx 86\%$ accuracy. Although accuracy marginally further increases with 8192 points, we limited the number of points to 2048 in favour of computational efficiency.}
\label{figure:resolutionExp}
\end{figure}

While investigating the impact of varying point cloud resolutions on the \acrshort{pointnet} model's accuracy, we observed a gradual decrease in \acrshort{kNN} accuracy as we decreased the number of points uniformly, using the farthest point sampling algorithm \cite{Eldar1994} as discussed in section \ref{datasetprep}. As shown in Fig. \ref{figure:resolutionExp}a, when using $50\%$ of the total 2048 points, the accuracy dropped by approximately $0.17\%$. When the number of input points is just 64, the accuracy only drops by  $\approx 13\%$, demonstrating the robustness of the \acrshort{pointnet} architecture to missing data and outliers. The phenomenon called ``depth discrepancies" naturally occurs in the wild due to the sensitivity of depth cameras to lighting conditions, as shown in Fig. \ref{figure:resolutionExp}b. This aspect is worth noting because point clouds are much sparser than depth maps. In this study, we process the depth maps at a raster resolution of (244x244) or 50,176 (u, v, z) points, compared to just 2048 (x, y, z) points, allowing faster inferencing with the \acrshort{pointnet} architecture. Although the accuracy increases with 8192 points, the increment is marginal. Therefore, we limited the number of input points to 2048, favouring the computation efficiency.

\textbf{Failure cases.} Although \acrshort{rgb} models give the best \acrshort{kNN} accuracy, the difference compared to the depth counterparts is negligible. As such, the \acrshort{resnet}-50 (Depth) and \acrshort{pointnet} provide state-of-the-art identification performance using depth data, which means that, in principle, they should work on other breeds that do not have coat patterns.

\begin{figure}[!b]
\centering
\includegraphics[width=1\textwidth]{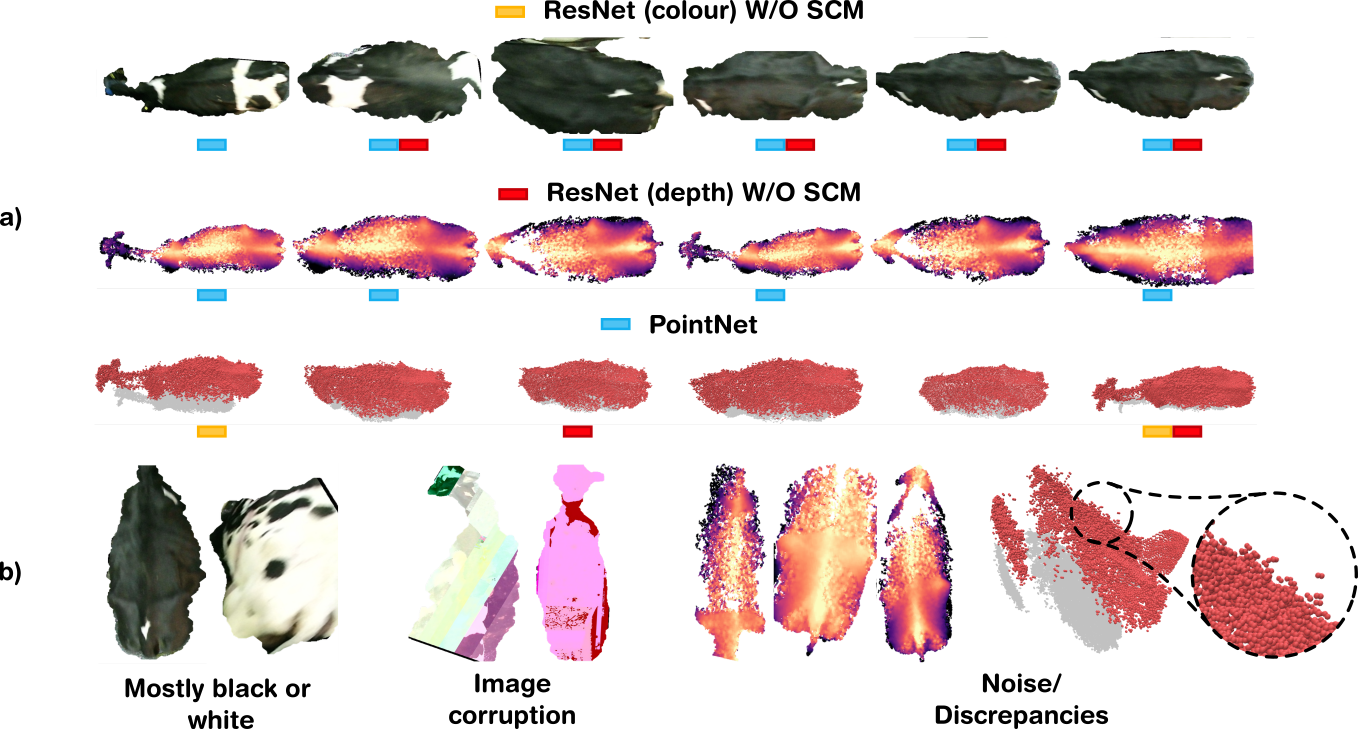}
\caption{\textbf{Identification failure cases.} \textbf{a)} The failure cases for \acrshort{resnet}-50 with and without the \acrshort{scm} model are shown together. The models that correctly identified the failure cases are indicated by the coloured rectangles. \textbf{b)} The \acrshort{rgb} failures are either corrupted or do not have a distinctive coat pattern. Other models failed due to noisy depth maps or point clouds.}
\label{figure:failures}
\end{figure}

Fig. \ref{figure:failures} illustrates the failure cases for the three input modalities, \acrshort{rgb} images, depth maps and point clouds. We observe that the \acrshort{resnet}-50 colour models with and without \acrshort{scm} fail when the input images do not have a distinctive coat pattern, such as primarily black or white cows. It is worth noting that when the colour models fail due to the lack of coat patterns, the \acrshort{resnet}-50, depth, and the \acrshort{pointnet} models are successful. The primary failure mode of the \acrshort{resnet}-50 (depth, w/o \acrshort{scm}) and \acrshort{pointnet} is noise.

\begin{figure}[!b]
\centering
\includegraphics[width=1\textwidth]{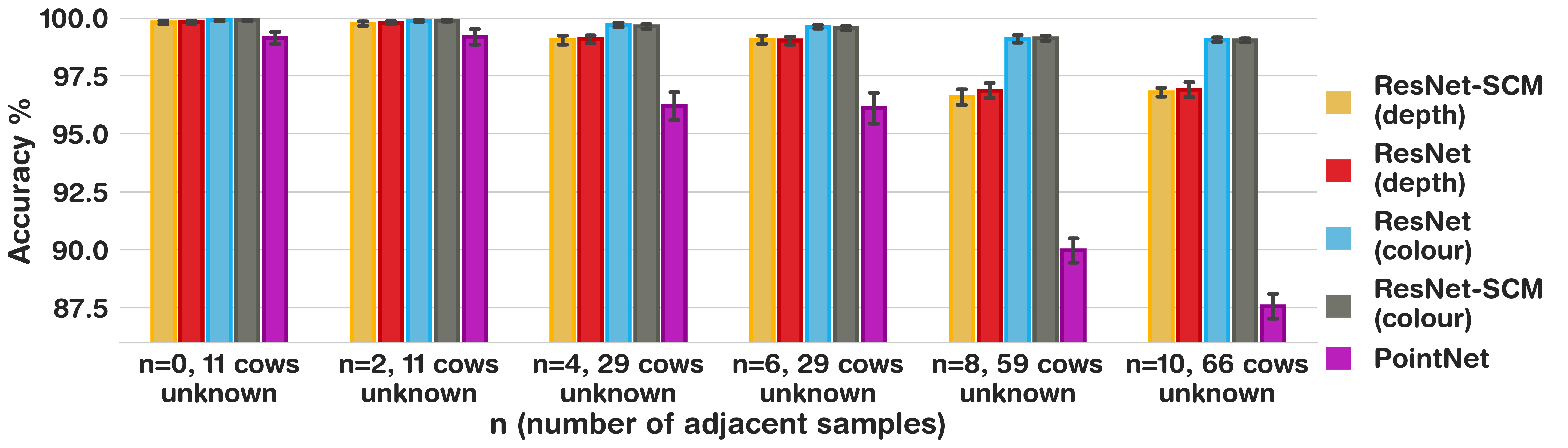}
\caption{\textbf{Open-set testing by leave-out sequence training.} n is the number of adjacent images to each test image removed from the training dataset. As n increases, the number of unknowns increases due to the lack of training data.}
\label{figure:nvsaccuracy}
\end{figure}

\textbf{Open-set evaluation.} The high accuracies of the models in Table \ref{tab:closedset} can be attributed to the sequential nature of the dataset. Therefore, it is essential to study the performance of the models without temporal bias. As mentioned in the section \ref{section:datasetprep}, we randomly split the dataset at a ratio of 7:3, five times and then removed n samples from the train set adjacent to the test data. Consequently, the process eliminates all training samples of some animals, which are considered `unknowns' and part of the open-set. The distinction between close and open-set training is twofold: \textbf{1)} We train the models only on the known animals, and \textbf{2)} For evaluation, we first generate embeddings for all knowns and unknowns and fit the \acrshort{kNN} classifier with the corresponding labels in the training set. Since we do not have any occurrences of unknowns, the training data, prior to temporal bias removal, is utilised. Therefore, the classifier can predict known individuals and enrol new individuals who were unknown during the training of the models. Finally, as shown in Fig. \ref{figure:nvsaccuracy}, we report the accuracy of the classifier on the test set comprised of knowns and unknowns, averaged over the five random splits.

As can be observed in Fig. \ref{figure:nvsaccuracy}, the performance difference between \acrshort{resnet}-50 w/o the \acrshort{scm} layer using \acrshort{rgb} coat patterns is negligible $(M=0.03, SD=0.15)$, regardless of the number n. As expected, a gradual decrease in performance is seen as n increases, indicating that the raw training set covers most of the variance in the test set, allowing the model to achieve high accuracy in the previous experiment (see Table \ref{tab:closedset}). Similarly, the depth-based \acrshort{resnet} variants (w/o \acrshort{scm}) also show a decreasing trend with n, although the performance is slightly worse than the models trained on coat pattern images.

Of all models, the \acrshort{pointnet} consistently performed worst (min: $86.1\%$; n=10, max: $99.36\%$; n=0). There are two reasons that influence the performance of \acrshort{pointnet}: \textbf{1)} As seen in Fig. \ref{figure:resolutionExp}, the accuracy depends on the density of the input point cloud, in-fact when we increased the number of input points to 8192 and retrained the \acrshort{pointnet} model for n=10, the minimum accuracy increased to $89.72\%$ from $86.1\%$, and \textbf{2)} For all \acrshort{resnet} experiments, we followed the standard practice of using the publically available weights, which were trained for a thousand classes of the ImageNet dataset. In contrast, due to the lack of large point cloud datasets, we pre-trained our \acrshort{pointnet} models on Modelnet40, which has just 40 classes. Therefore, the two different architectures can not be compared solely based on the open-set accuracy.

\begin{figure}[!b]
\centering
\includegraphics[width=1\textwidth]{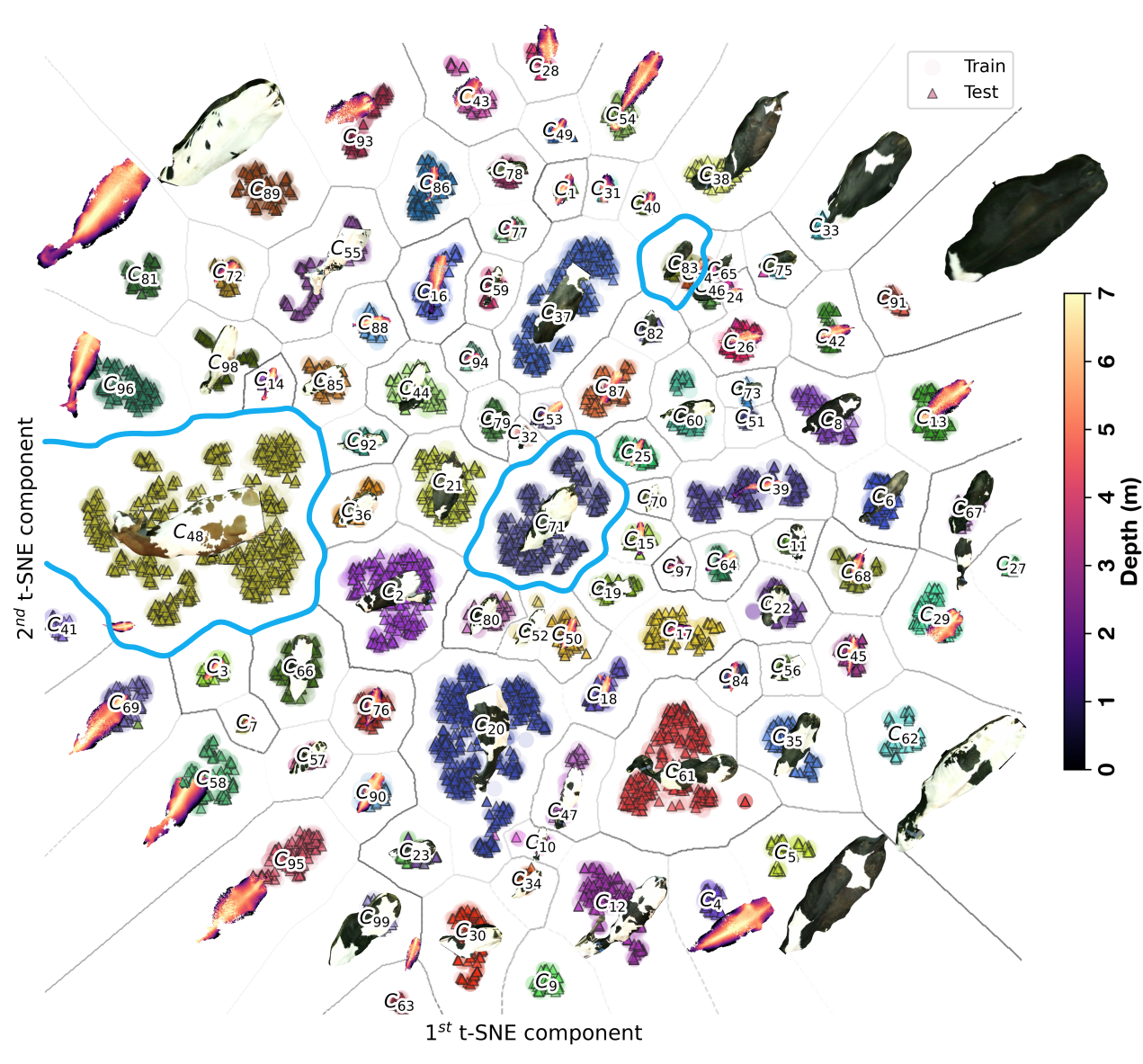}
\caption{\textbf{\acrshort{tsne} visualisation of \acrshort{resnet}-50 depth embeddings for CowDepth2023.} The high-dimensional latent space is projected down to 2-D space. We cluster the training embeddings into ID groups as shown in colour and overlay the test embeddings on top. Testing and training examples are coincident due to classification success. The \acrshort{resnet} model consumes a depth map, not a colour image of the coat patterns. The coat patterns next to the cluster labels have been swapped randomly with the input depth maps with a probability of 0.7 for visualisation only.}
\label{figure:tsne-resnet}
\end{figure}

\textbf{Embedding visualisations using \acrfull{tsne}.} Figure \ref{figure:tsne-resnet} depicts the training and testing embeddings projected onto a two-dimensional space for all the examples in the dataset, respectively. The top performing depth-based \acrshort{resnet}-50 exhibits the desired property of having a discriminative embedding space.

To visualise the \acrshort{kNN} decision boundaries and create a consistent latent space, we reduced dimensionality on the 128-dimensional embeddings obtained from the test-train split using \acrshort{tsne}. Subsequently, a \acrshort{kNN} classifier was applied to the reduced training embeddings. The \acrshort{kNN} classifier applied to the reduced 2D \acrshort{tsne} output was solely utilised for visualisation purposes, while the actual labelling was conducted independently on the complete 128-dimensional embeddings.

\begin{figure}[!b]
\centering
\includegraphics[width=1\textwidth]{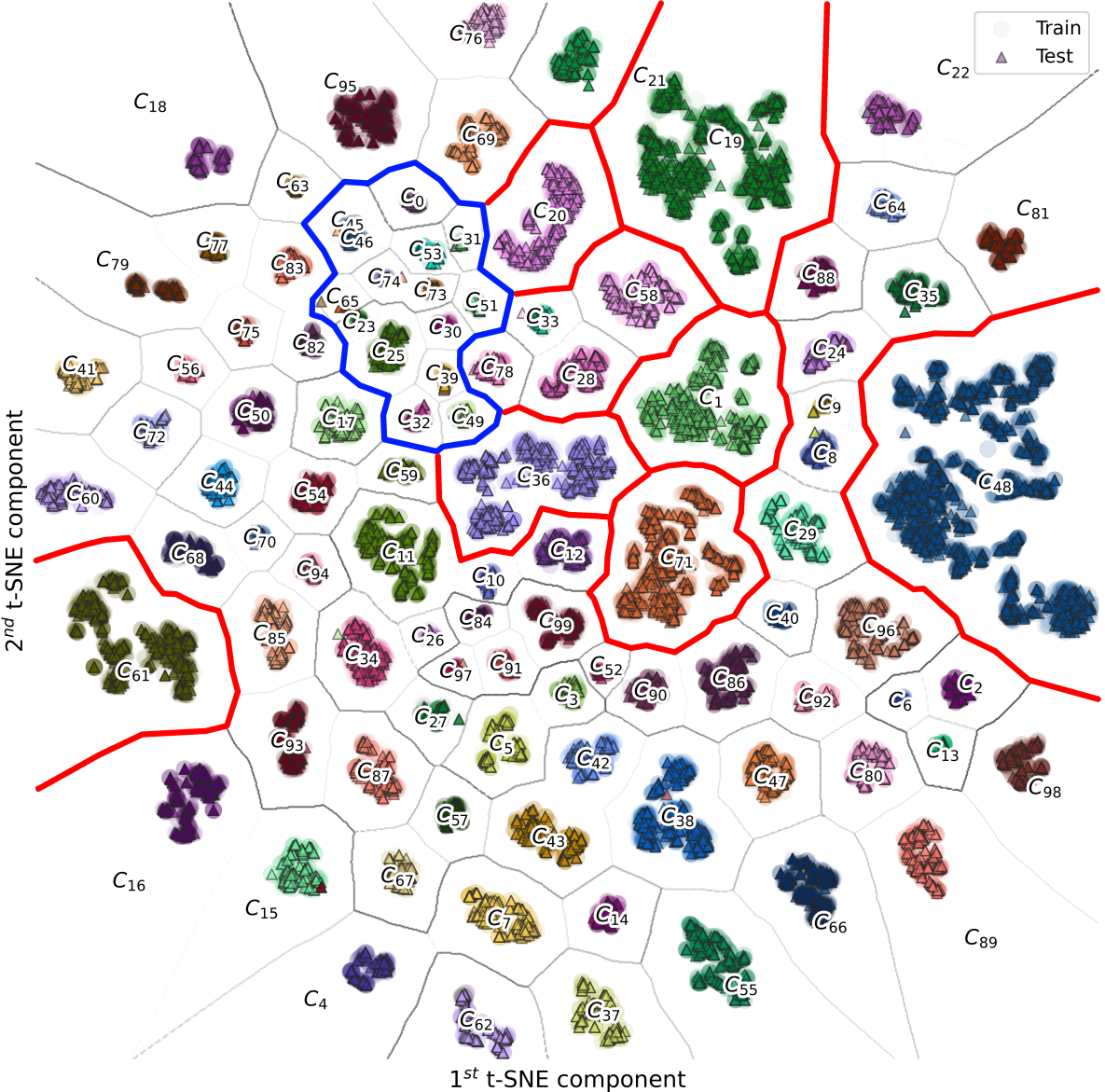}
\caption{\textbf{\acrshort{tsne} visualisation for \acrshort{pointnet}} Shows \acrshort{tsne} reduced embeddings of individual cows ($C_0\cdots C_{99}$). To visualise the \acrshort{kNN} decision boundaries (in dashed black), the 128-component embeddings for the test and train set were simultaneously reduced to two components for a consistent latent space. A \acrshort{kNN} was then fitted on the reduced embeddings for the training set.}
\label{figure:pointnetTsne}
\end{figure}

In Fig. \ref{figure:pointnetTsne}, we present the kNN decision boundaries in conjunction with the \acrshort{tsne}-reduced \acrshort{pointnet} embeddings. Some embeddings form compact, well-separated, but densely populated clusters (see Fig. \ref{figure:pointnetTsne}, top-left, highlighted in blue).
In such clusters, the apparent decision boundaries suggest that the \acrshort{pointnet} model is sufficient in learning some subtle, distinctive patterns. Furthermore, cows such as $C_{48}$, $C_{61}$, and $C_{71}$ form well-separated but spread out clusters, which seems to indicate high intra-class variance in the latent space (see Fig. \ref{figure:pointnetTsne}, highlighted in red). However, the minority of such clusters (7 out of 100) implies the model's efficacy for the identification task. Lastly, it is important to note that both phenomena are also present in the case of \acrshort{resnet}-50 depth, see $C_{48}$, $C_{71}$ for large clusters and $C_{83}$ for compact clusters in Fig. \ref{figure:tsne-resnet} (highlighted in light blue).

To further understand the efficacy of our models on the variety of input modalities,  we utilised two visualisation techniques, \acrfull{gradcam} and \acrfull{pcsm}, that rely on the outputs from intermediate layers of the networks to spatially rank the importance of the regions in the input data for \acrshort{resnet} and \acrshort{pointnet} architectures, respectively.

\textbf{\acrshort{gradcam}s and \acrshort{pcsm}s.} We modified the \acrshort{resnet}s to output the network gradients during a backward pass. Then, using a randomly chosen depth map or colour image from the CowDepth2023 dataset, we performed a forward pass through the network. Since the networks require three images, we passed the same image duplicated three times into the network. Unlike prior work, such as \citet{chen2020adapting}, where the class probabilities are calculated from the embeddings, we found that utilising the softmax layer output from the network to calculate the loss (i.e. $\mathbb{L}_{SoftMax}$) was sufficient. After the back-propagation, the importance of spatial locations, $w^c$, can be obtained by analysing the gradients for neurons $y^c$ that contributed most in predicting the probability of a class $c$ with respect to the activations of the final convolutional layer $A$  \cite{selvaraju2017grad}. 

$$w^c = AvgPool (\frac{\delta y^c}{\delta A})$$

The heatmap $H$ depicting the spatial importance can be obtained by linearly combining average-pooled gradients $w^c$ with the activations $A$, passed through a ReLU function. Finally, The resulting heatmap $H$ was superimposed onto the original depth image to produce the visualisations.

In the context of the \acrshort{pointnet} architecture, the max-pooling layer contributes towards the embedding by filtering out the least essential points in the point cloud across all the channels of output from the previous \acrfull{mlp} layer. Although the output from the max-pooling layer can be directly utilised to retrieve the indices of the points that are critical to a successful identification, we used a gradient-based technique as described by \citet{Zheng2019}. First, the input 3D points $x_{i} \in X$ are converted from the cartesian coordinates $(x,y,z)$ to spherical coordinates $x_{c}$, $(r, \psi, \phi)$ in order to make the gradient calculations invariant to orientation. Next, a certain amount of points (ten, in our case) are dropped randomly from the point cloud until the model wrongly classifies the point cloud. Similar to the \acrshort{gradcam}, the importance of points is measured by computing the gradient of the softmax loss $\mathbb{L}_{SoftMax}$ w.r.t. $r$ as follows:

$$\frac{\partial \mathbb{L}_{SoftMax}}{\partial r_i} = \sum_{j=1}^{3} \frac{\partial \mathbb{L}_{SoftMax}}{\partial x_{ij}} \frac{x_{ij}-x_{cj}}{r_i}$$

Then, the saliency score for all the points can be calculated as:

$$s_i = -\frac{\partial \mathbb{L}_{SoftMax}}{\partial r_i}r_i$$

\begin{figure}[!b]
\centering
\includegraphics[width=1\textwidth]{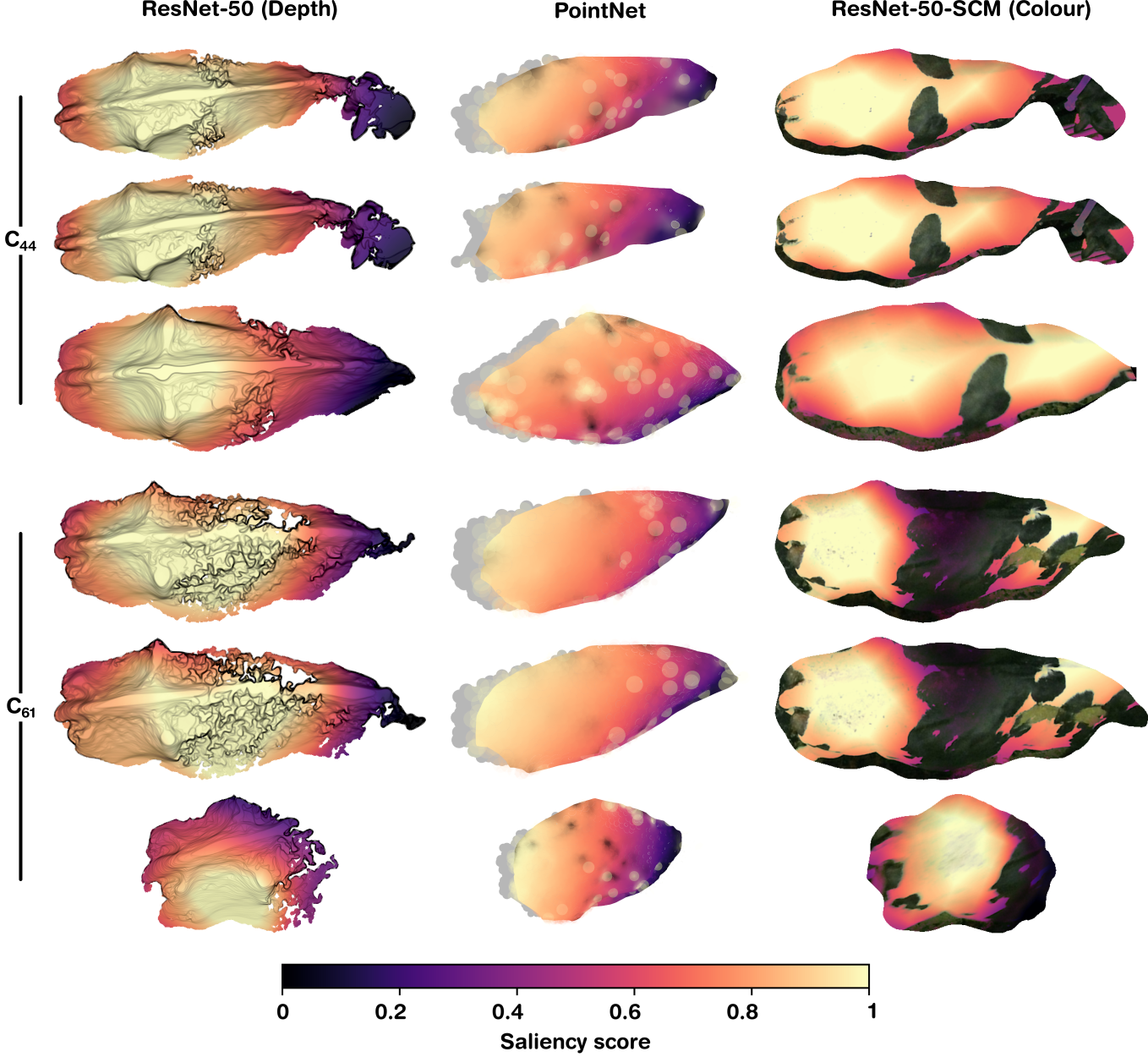}
\caption{\textbf{Region importance visualisation.} Lighter colours correspond to regions that the \acrshort{pointnet} and \acrshort{resnet} models (left to right) deem most important for the two individuals. It is important to note that we colour-coded the point clouds by the normalised rank of saliency scores $s_i$. Furthermore, the point clouds appear sparse compared to the depth maps because they only consist of 2048 points.}
\label{figure:pointnetGradcam}
\end{figure}

Fig. \ref{figure:pointnetGradcam} shows the \acrshort{gradcam} and \acrshort{pcsm} visualisations for the top performing models on the open-set experiments. We observe that for both cows $C_{61}$ and $C_{44}$, which correspond to highly spread out clusters in Fig. \ref{figure:pointnetTsne} and Fig. \ref{figure:tsne-resnet}. Despite the sizeable intra-class variance in the two clusters, the \acrshort{resnet}-50 (depth) and \acrshort{pointnet} architectures both rely on bone structures such as the dorsal vertebra and the thurl-hook-pin region. At the same time, the neck region is of the least spatial importance. Furthermore, we observe that the \acrshort{resnet}-50 (Colour) model tends to focus more on the white markings, as evidenced by the fact that black regions appear dark blue (see Fig. \ref{figure:pointnetGradcam}, $C_{61}$). Interestingly, both PointNet and ResNet models focus on similar areas, indicating a certain level of consistency in their decision-making process. A shared agreement between the models suggests that these regions are universally crucial for the classification task regardless of the architecture. Furthermore, the lack of focus on the neck indicates that it is challenging to extract discriminative features, possibly because of significant noise in the 3D data (see section \ref{section:datasetprep}, discussion on median filter and depth discrepancies). The other potential cause could be the physical movement of the neck, which forces the model to focus on more stable regions. It is essential to highlight that the above findings hold despite the augmentations we used during the \acrshort{resnet} training. Finally, \acrshort{gradcam} heatmaps of colour-based \acrshort{resnet} architecture suggest that the model would fail when the coat patterns primarily consist of black or white regions, as we highlighted during the discussion about failure cases (see Fig. \ref{figure:failures}b).

\section{Conclusion}
\label{Conclusion}
Our primary objective throughout project execution has been to extend our previous work on identifying individual cows using RGB imagery to produce \acrshort{dnn}s and \acrshort{mlp}s capable of identifying cattle through depth imagery. Our motivation behind the depth map and point cloud approach was to enable the identification of individuals who were primarily black or white for whom the coat pattern-based approach fell short (as shown in Fig. \ref{figure:failures}). Furthermore, through the process of open-set evaluation, we demonstrated that our methodology does not rely on network retraining for enrolment or removal of individuals. Finally, we qualitatively assessed the efficacy of depth maps and point clouds for cattle identification by visualising the gradients of \acrshort{resnet} and \acrshort{pointnet} architectures through \acrshort{gradcam} and \acrshort{pcsm}, respectively.

To our knowledge, depth as a modality for cattle identification has yet to be studied. Therefore, we presented a new dataset, CowDepth2023, to facilitate experiments with the two architectures and allow detailed discussion on their efficacy.

We compared the performance of the \acrshort{pointnet} and \acrshort{resnet} architectures. Notably, both the \acrshort{pointnet} and the \acrshort{resnet} (depth, w/o \acrshort{scm}) models performed on par with the colour baselines. Adding a spatial context module before the fully connected layers in \acrshort{resnet} yielded mixed results. In the case of coat patterns, the \acrshort{scm} layer improved the accuracy, whereas \acrshort{resnet} without the \acrshort{scm} performed better. We assessed the robustness of the PointNet architecture by varying the point cloud resolution, i.e. the number of points. Despite uniform resampling and a gradual decrease in points, the PointNet model exhibited robustness to missing data and outliers. The accuracy dropped only by $13\%$ when there were just 64 points.

We visualised the embeddings of the models using \acrshort{tsne}, revealing interesting cluster patterns. Some clusters showed compactness and separation, while others were spread out, indicating intra-class variance. Both \acrshort{pointnet} and \acrshort{resnet} models exhibited similar cluster patterns, suggesting a consistent latent space. To understand the factors influencing the cluster patterns, we employed \acrshort{gradcam} and \acrshort{pcsm} techniques to rank the importance of regions in the input data. Both methods revealed the significance of bony regions and the relative lack of importance given to the neck region. The consistency in region importance between the two architectures adds evidence that these are universally critical features for classification.

Overall, this study provides valuable insights into effective cattle identification using depth data via deep metric learning and opens up avenues for further research in this domain.
\section{Future work}
The results from this research have demonstrated that there is scope for deep learning applications to use depth imagery alone to classify individual cattle in a herd. The accuracy achieved by the proposed network suggests that it could be a viable solution for improving efficiency in the cattle farming industry. However, further research is needed to fully evaluate the practicalities and extensibility of the approach to real-world applications.

\textbf{Variation of input imagery.} Extending the technique to handle footage from any angle would be necessary for our methodology to be scalable in large farming operations. It is also not clear from our experiments how well the system would hold up to variation in input depth imagery over time. For example, an interesting avenue would be to train a model using data containing images of cows in different situations, orientations, and locations rather than from a single sequential image stream.

Additionally, in this study, we focused on the Holstein Friesian breed specifically because we wanted to compare the depth map based approach with \acrshort{rgb}. Therefore, a natural extension would be to test our system on truly patternless breeds. Furthermore, our work can also be translated to other livestock species, such as sheep, goats and pigs.

As evident from the depth maps in Fig. \ref{figure:pointnetGradcam}, the Kinect V2 camera struggles to measure distances in black portions of the coat pattern. The noise occurs for multiple reasons, such as specular highlights. However, the most likely cause is the black surfaces, which absorb the near-infrared spectrum that the time-of-flight cameras such as the Kinect V2 rely on for deriving accurate measurements \cite{Haider2022}. Despite the measurement noise, both models (\acrshort{resnet} (depth) and \acrshort{pointnet}) achieve high levels of accuracy by focusing on regions where the phenomenon is less apparent. The future extension to our methodology could be to study the noise characteristics of other 3D camera technologies, such as stereo depth and structured light.

The new data collection process would also allow us to study how identification accuracy drifts over time as body condition changes due to exogenous factors. As our study highlights the consistency of particular regions that the models choose to focus on, the new data collection would also allow us to consider factors such as age, breed variations, health conditions, and environmental factors such as diet or living conditions, potentially providing simultaneous prediction for key production and welfare indicators. 

\textbf{Combining data sources.} It could also be beneficial to explore whether adding other forms of input data would benefit the system. For example, while discussing the failure cases, we observed that depth-based networks failed due to noise, while \acrshort{rgb} failed on predominantly black or white cows. Developing a network where predictions are made by combining \acrshort{rgb} and depth map/point cloud inputs could result in a more robust design.

In a highly instrumented farm setting, it could even be helpful to explore whether additional remote sensors, such as thermography, could provide more contextual information about the cows and their environment, allowing further insights into the well-being and body condition of each cow (see review by \citet{McManus2022}). Extending the identification system in such ways would represent a significant advancement towards automating each animal's welfare assessments.

\textbf{Detecting new clusters.} Currently, our approach requires prior knowledge of the number of individuals for fitting the \acrshort{kNN} algorithm. As proposed by \citet{Gao2022}, a scalable labelling system to annotate new individuals can serve as a practical approach. However, the problem of detecting clusters of new individuals from a static embedding space presents an interesting avenue for fundamental research.

\section{Acknowledgements}
This study is supported by a studentship from the Biotechnology and Biological Sciences Research Council (BBSRC; UK) South West Biosciences Doctoral Training Programme (Project ref. \href{https://gtr.ukri.org/projects?ref=studentship-2593504}{2593504}, \href{https://gtr.ukri.org/projects?ref=BB/T008741/1}{BB/T008741/1}).

\textbf{Declaration of generative AI and AI-assisted technologies in the writing process.} During the preparation of this work, the author(s) used \href{https://app.grammarly.com/}{Grammarly} in order to improve the readability and language of the work and \textbf{not} to replace key authoring tasks such as producing scientific, pedagogic, or medical insights, drawing scientific conclusions, or providing clinical recommendations. After using \href{https://app.grammarly.com/}{Grammarly}, the author(s) reviewed and edited the content as needed and take(s) full responsibility for the content of the publication.
\bibliographystyle{elsarticle-num-names} 
\bibliography{elsarticle-template-num}

\begin{thebibliography}{79}
\expandafter\ifx\csname natexlab\endcsname\relax\def\natexlab#1{#1}\fi
\providecommand{\url}[1]{\texttt{#1}}
\providecommand{\href}[2]{#2}
\providecommand{\path}[1]{#1}
\providecommand{\DOIprefix}{doi:}
\providecommand{\ArXivprefix}{arXiv:}
\providecommand{\URLprefix}{URL: }
\providecommand{\Pubmedprefix}{pmid:}
\providecommand{\doi}[1]{\href{http://dx.doi.org/#1}{\path{#1}}}
\providecommand{\Pubmed}[1]{\href{pmid:#1}{\path{#1}}}
\providecommand{\bibinfo}[2]{#2}
\ifx\xfnm\relax \def\xfnm[#1]{\unskip,\space#1}\fi
\bibitem[{McAuliffe et~al.(2018)McAuliffe, Takahashi, Orr, Harris, and Lee}]{McAuliffe2018}
\bibinfo{author}{G.~A. McAuliffe}, \bibinfo{author}{T.~Takahashi}, \bibinfo{author}{R.~J. Orr}, \bibinfo{author}{P.~Harris}, \bibinfo{author}{M.~R. Lee},
\newblock \bibinfo{title}{Distributions of emissions intensity for individual beef cattle reared on pasture-based production systems},
\newblock \bibinfo{journal}{Journal of Cleaner Production} \bibinfo{volume}{171} (\bibinfo{year}{2018}) \bibinfo{pages}{1672--1680}. \DOIprefix\doi{10.1016/J.JCLEPRO.2017.10.113}.
\bibitem[{Sumner et~al.(2018)Sumner, von Keyserlingk, and Weary}]{Sumner2018}
\bibinfo{author}{C.~L. Sumner}, \bibinfo{author}{M.~A. von Keyserlingk}, \bibinfo{author}{D.~M. Weary},
\newblock \bibinfo{title}{Perspectives of farmers and veterinarians concerning dairy cattle welfare},
\newblock \bibinfo{journal}{Animal frontiers : the review magazine of animal agriculture} \bibinfo{volume}{8} (\bibinfo{year}{2018}) \bibinfo{pages}{8--13}. \URLprefix \url{https://pubmed.ncbi.nlm.nih.gov/32002209/}. \DOIprefix\doi{10.1093/AF/VFX006}.
\bibitem[{DeGraves and Fetrow(1993)}]{DeGraves1993}
\bibinfo{author}{F.~J. DeGraves}, \bibinfo{author}{J.~Fetrow},
\newblock \bibinfo{title}{Economics of mastitis and mastitis control},
\newblock \bibinfo{journal}{The Veterinary clinics of North America. Food animal practice} \bibinfo{volume}{9} (\bibinfo{year}{1993}) \bibinfo{pages}{421--434}. \URLprefix \url{https://pubmed.ncbi.nlm.nih.gov/8242449/}. \DOIprefix\doi{10.1016/S0749-0720(15)30611-3}.
\bibitem[{Shahbaz et~al.(2024)Shahbaz, Zhang, and Smith}]{Shahbaz2024}
\bibinfo{author}{A.~Shahbaz}, \bibinfo{author}{W.~Zhang}, \bibinfo{author}{M.~Smith},
\newblock \bibinfo{title}{A two-stage approach using yolo for automated assessment of digital dermatitis within dairy cattle},
\newblock in: \bibinfo{booktitle}{2024 IEEE 22nd World Symposium on Applied Machine Intelligence and Informatics (SAMI)}, \bibinfo{publisher}{Institute of Electrical and Electronics Engineers (IEEE)}, \bibinfo{year}{2024}, pp. \bibinfo{pages}{000417--000422}. \DOIprefix\doi{10.1109/SAMI60510.2024.10432745}.
\bibitem[{Disney et~al.(2001)Disney, Green, Forsythe, Wiemers, and Weber}]{Disney2001}
\bibinfo{author}{W.~T. Disney}, \bibinfo{author}{J.~W. Green}, \bibinfo{author}{K.~W. Forsythe}, \bibinfo{author}{J.~F. Wiemers}, \bibinfo{author}{S.~Weber},
\newblock \bibinfo{title}{Benefit-cost analysis of animal identification for disease prevention and control},
\newblock \bibinfo{journal}{Revue scientifique et technique (International Office of Epizootics)} \bibinfo{volume}{20} (\bibinfo{year}{2001}) \bibinfo{pages}{385--405}. \URLprefix \url{https://pubmed.ncbi.nlm.nih.gov/11552703/}. \DOIprefix\doi{10.20506/RST.20.2.1277}.
\bibitem[{Andrew et~al.(2021)Andrew, Gao, Mullan, Campbell, Dowsey, and Burghardt}]{Andrew2021}
\bibinfo{author}{W.~Andrew}, \bibinfo{author}{J.~Gao}, \bibinfo{author}{S.~Mullan}, \bibinfo{author}{N.~Campbell}, \bibinfo{author}{A.~W. Dowsey}, \bibinfo{author}{T.~Burghardt},
\newblock \bibinfo{title}{Visual identification of individual holstein-friesian cattle via deep metric learning},
\newblock \bibinfo{journal}{Computers and Electronics in Agriculture} \bibinfo{volume}{185} (\bibinfo{year}{2021}) \bibinfo{pages}{106133}. \DOIprefix\doi{10.1016/J.COMPAG.2021.106133}.
\bibitem[{{Department for Environment, Food and Rural Affairs}(2008)}]{DEFRA2008}
\bibinfo{author}{{Department for Environment, Food and Rural Affairs}}, \bibinfo{title}{The cattle book 2008 - gov.uk}, \bibinfo{year}{2008}. \URLprefix \url{https://www.gov.uk/government/statistics/the-cattle-book-2008}.
\bibitem[{Bezen et~al.(2020)Bezen, Edan, and Halachmi}]{Bezen2020}
\bibinfo{author}{R.~Bezen}, \bibinfo{author}{Y.~Edan}, \bibinfo{author}{I.~Halachmi},
\newblock \bibinfo{title}{Computer vision system for measuring individual cow feed intake using rgb-d camera and deep learning algorithms},
\newblock \bibinfo{journal}{Computers and Electronics in Agriculture} \bibinfo{volume}{172} (\bibinfo{year}{2020}) \bibinfo{pages}{105345}. \DOIprefix\doi{10.1016/J.COMPAG.2020.105345}.
\bibitem[{Okura et~al.(2019)Okura, Ikuma, Makihara, Muramatsu, Nakada, and Yagi}]{Okura2019}
\bibinfo{author}{F.~Okura}, \bibinfo{author}{S.~Ikuma}, \bibinfo{author}{Y.~Makihara}, \bibinfo{author}{D.~Muramatsu}, \bibinfo{author}{K.~Nakada}, \bibinfo{author}{Y.~Yagi},
\newblock \bibinfo{title}{Rgb-d video-based individual identification of dairy cows using gait and texture analyses},
\newblock \bibinfo{journal}{Computers and Electronics in Agriculture} \bibinfo{volume}{165} (\bibinfo{year}{2019}) \bibinfo{pages}{104944}. \DOIprefix\doi{10.1016/J.COMPAG.2019.104944}.
\bibitem[{Stankovski et~al.(2012)Stankovski, Ostojic, Senk, Rakic-Skokovic, Trivunovic, and Kucevic}]{Stankovski2012}
\bibinfo{author}{S.~Stankovski}, \bibinfo{author}{G.~Ostojic}, \bibinfo{author}{I.~Senk}, \bibinfo{author}{M.~Rakic-Skokovic}, \bibinfo{author}{S.~Trivunovic}, \bibinfo{author}{D.~Kucevic},
\newblock \bibinfo{title}{Dairy cow monitoring by rfid},
\newblock \bibinfo{journal}{Scientia Agricola} \bibinfo{volume}{69} (\bibinfo{year}{2012}) \bibinfo{pages}{75--80}. \URLprefix \url{https://www.scielo.br/j/sa/a/jtc43hq5QNh8cXnW4WVLRmF/?lang=en}. \DOIprefix\doi{10.1590/S0103-90162012000100011}.
\bibitem[{Adam et~al.(2016)Adam, Holcomb, Buser, Mayfield, and Thomas}]{Adam2016}
\bibinfo{author}{B.~D. Adam}, \bibinfo{author}{R.~Holcomb}, \bibinfo{author}{M.~Buser}, \bibinfo{author}{B.~Mayfield}, \bibinfo{author}{J.~Thomas},
\newblock \bibinfo{title}{Enhancing food safety, product quality, and value-added in food supply chains using whole-chain traceability},
\newblock \bibinfo{journal}{International Food and Agribusiness Management Review} \bibinfo{volume}{19} (\bibinfo{year}{2016}).
\bibitem[{Awad(2016)}]{Awad2016}
\bibinfo{author}{A.~I. Awad},
\newblock \bibinfo{title}{From classical methods to animal biometrics: A review on cattle identification and tracking},
\newblock \bibinfo{journal}{Computers and Electronics in Agriculture} \bibinfo{volume}{123} (\bibinfo{year}{2016}) \bibinfo{pages}{423--435}. \DOIprefix\doi{10.1016/J.COMPAG.2016.03.014}.
\bibitem[{Johnston and Edwards(1996)}]{Johnston1996}
\bibinfo{author}{A.~M. Johnston}, \bibinfo{author}{D.~S. Edwards},
\newblock \bibinfo{title}{Welfare implications of identification of cattle by ear tags},
\newblock \bibinfo{journal}{Veterinary Record} \bibinfo{volume}{138} (\bibinfo{year}{1996}) \bibinfo{pages}{612--614}. \URLprefix \url{https://onlinelibrary.wiley.com/doi/full/10.1136/vr.138.25.612 https://onlinelibrary.wiley.com/doi/abs/10.1136/vr.138.25.612 https://bvajournals.onlinelibrary.wiley.com/doi/10.1136/vr.138.25.612}. \DOIprefix\doi{10.1136/VR.138.25.612}.
\bibitem[{Chapa et~al.(2020)Chapa, Maschat, Iwersen, Baumgartner, and Drillich}]{Chapa2020}
\bibinfo{author}{J.~M. Chapa}, \bibinfo{author}{K.~Maschat}, \bibinfo{author}{M.~Iwersen}, \bibinfo{author}{J.~Baumgartner}, \bibinfo{author}{M.~Drillich},
\newblock \bibinfo{title}{Accelerometer systems as tools for health and welfare assessment in cattle and pigs – a review},
\newblock \bibinfo{journal}{Behavioural Processes} \bibinfo{volume}{181} (\bibinfo{year}{2020}) \bibinfo{pages}{104262}. \URLprefix \url{https://www.sciencedirect.com/science/article/pii/S0376635720304551}. \DOIprefix\doi{https://doi.org/10.1016/j.beproc.2020.104262}.
\bibitem[{Lind and Lindahl(2018)}]{Lind2018}
\bibinfo{author}{A.-K. Lind}, \bibinfo{author}{C.~Lindahl}, \bibinfo{title}{Moocall – en sensor med koll p{\aa} kalvningar}, \bibinfo{type}{Technical Report} \bibinfo{number}{2018:75}, RISE - Research Institutes of Sweden, Agrifood and Bioscience, \bibinfo{year}{2018}. \URLprefix \url{http://www.diva-portal.org/smash/record.jsf?pid=diva2\%3A1272143&dswid=-8189}.
\bibitem[{Ledgerwood et~al.(2010)Ledgerwood, Winckler, and Tucker}]{Ledgerwood2010}
\bibinfo{author}{D.~N. Ledgerwood}, \bibinfo{author}{C.~Winckler}, \bibinfo{author}{C.~B. Tucker},
\newblock \bibinfo{title}{Evaluation of data loggers, sampling intervals, and editing techniques for measuring the lying behavior of dairy cattle},
\newblock \bibinfo{journal}{Journal of Dairy Science} \bibinfo{volume}{93} (\bibinfo{year}{2010}) \bibinfo{pages}{5129--5139}. \DOIprefix\doi{10.3168/JDS.2009-2945}.
\bibitem[{Kokin et~al.(2014)Kokin, Praks, Veermäe, Poikalainen, and Vallas}]{Kokin2014}
\bibinfo{author}{E.~Kokin}, \bibinfo{author}{J.~Praks}, \bibinfo{author}{I.~Veermäe}, \bibinfo{author}{V.~Poikalainen}, \bibinfo{author}{M.~Vallas},
\newblock \bibinfo{title}{Icetag3d™ accelerometric device in cattle lameness detection},
\newblock \bibinfo{journal}{Agronomy Research} \bibinfo{volume}{12} (\bibinfo{year}{2014}).
\bibitem[{Li et~al.(2022)Li, Lei, and Liu}]{Li20222}
\bibinfo{author}{Z.~Li}, \bibinfo{author}{X.~Lei}, \bibinfo{author}{S.~Liu},
\newblock \bibinfo{title}{A lightweight deep learning model for cattle face recognition},
\newblock \bibinfo{journal}{Computers and Electronics in Agriculture} \bibinfo{volume}{195} (\bibinfo{year}{2022}) \bibinfo{pages}{106848}. \DOIprefix\doi{10.1016/J.COMPAG.2022.106848}.
\bibitem[{Kumar et~al.(2018)Kumar, Pandey, Satwik, Kumar, Singh, Singh, and Mohan}]{Kumar2018}
\bibinfo{author}{S.~Kumar}, \bibinfo{author}{A.~Pandey}, \bibinfo{author}{K.~S.~R. Satwik}, \bibinfo{author}{S.~Kumar}, \bibinfo{author}{S.~K. Singh}, \bibinfo{author}{A.~K. Singh}, \bibinfo{author}{A.~Mohan},
\newblock \bibinfo{title}{Deep learning framework for recognition of cattle using muzzle point image pattern},
\newblock \bibinfo{journal}{Measurement} \bibinfo{volume}{116} (\bibinfo{year}{2018}) \bibinfo{pages}{1--17}. \DOIprefix\doi{10.1016/J.MEASUREMENT.2017.10.064}.
\bibitem[{Lu et~al.(2014)Lu, He, Wen, and Wang}]{Lu2014}
\bibinfo{author}{Y.~Lu}, \bibinfo{author}{X.~He}, \bibinfo{author}{Y.~Wen}, \bibinfo{author}{P.~S. Wang},
\newblock \bibinfo{title}{A new cow identification system based on iris analysis and recognition},
\newblock \bibinfo{journal}{International Journal of Biometrics} \bibinfo{volume}{6} (\bibinfo{year}{2014}) \bibinfo{pages}{18--32}. \DOIprefix\doi{10.1504/IJBM.2014.059639}.
\bibitem[{Allen et~al.(2008)Allen, Golden, Taylor, Patterson, Henriksen, and Skuce}]{Allen2008}
\bibinfo{author}{A.~Allen}, \bibinfo{author}{B.~Golden}, \bibinfo{author}{M.~Taylor}, \bibinfo{author}{D.~Patterson}, \bibinfo{author}{D.~Henriksen}, \bibinfo{author}{R.~Skuce},
\newblock \bibinfo{title}{Evaluation of retinal imaging technology for the biometric identification of bovine animals in northern ireland},
\newblock \bibinfo{journal}{Livestock Science} \bibinfo{volume}{116} (\bibinfo{year}{2008}) \bibinfo{pages}{42--52}. \DOIprefix\doi{10.1016/J.LIVSCI.2007.08.018}.
\bibitem[{Eilertsen et~al.(2019)Eilertsen, Mantiuk, and Unger}]{Eilertsen2019}
\bibinfo{author}{G.~Eilertsen}, \bibinfo{author}{R.~K. Mantiuk}, \bibinfo{author}{J.~Unger},
\newblock \bibinfo{title}{Single-frame regularization for temporally stable cnns},
\newblock \bibinfo{journal}{Proceedings of the IEEE Computer Society Conference on Computer Vision and Pattern Recognition} \bibinfo{volume}{2019-June} (\bibinfo{year}{2019}) \bibinfo{pages}{11168--11177}. \URLprefix \url{http://arxiv.org/abs/1902.10424 http://dx.doi.org/10.1109/CVPR.2019.01143}. \DOIprefix\doi{10.1109/CVPR.2019.01143}.
\bibitem[{Xu et~al.(2022)Xu, Wang, Guo, Chen, Li, Cao, and Wu}]{Xu2022}
\bibinfo{author}{B.~Xu}, \bibinfo{author}{W.~Wang}, \bibinfo{author}{L.~Guo}, \bibinfo{author}{G.~Chen}, \bibinfo{author}{Y.~Li}, \bibinfo{author}{Z.~Cao}, \bibinfo{author}{S.~Wu},
\newblock \bibinfo{title}{Cattlefacenet: A cattle face identification approach based on retinaface and arcface loss},
\newblock \bibinfo{journal}{Computers and Electronics in Agriculture} \bibinfo{volume}{193} (\bibinfo{year}{2022}) \bibinfo{pages}{106675}. \DOIprefix\doi{10.1016/J.COMPAG.2021.106675}.
\bibitem[{Gong et~al.(2022)Gong, Pan, Chen, Hu, Li, Sun, Mu, and Guo}]{Gong2022}
\bibinfo{author}{H.~Gong}, \bibinfo{author}{H.~Pan}, \bibinfo{author}{L.~Chen}, \bibinfo{author}{T.~Hu}, \bibinfo{author}{S.~Li}, \bibinfo{author}{Y.~Sun}, \bibinfo{author}{Y.~Mu}, \bibinfo{author}{Y.~Guo},
\newblock \bibinfo{title}{Facial recognition of cattle based on sk-resnet},
\newblock \bibinfo{journal}{Scientific Programming} \bibinfo{volume}{2022} (\bibinfo{year}{2022}). \DOIprefix\doi{10.1155/2022/5773721}.
\bibitem[{Yao et~al.(2019)Yao, Liu, Hu, Kuang, Liu, and Gao}]{Yao2019}
\bibinfo{author}{L.~Yao}, \bibinfo{author}{H.~Liu}, \bibinfo{author}{Z.~Hu}, \bibinfo{author}{Y.~Kuang}, \bibinfo{author}{C.~Liu}, \bibinfo{author}{Y.~Gao},
\newblock \bibinfo{title}{Cow face detection and recognition based on automatic feature extraction algorithm},
\newblock \bibinfo{journal}{ACM International Conference Proceeding Series}  (\bibinfo{year}{2019}). \URLprefix \url{https://dl.acm.org/doi/10.1145/3321408.3322628}. \DOIprefix\doi{10.1145/3321408.3322628}.
\bibitem[{Weng et~al.(2022)Weng, Liu, Zheng, Zhang, and Gong}]{Weng2022}
\bibinfo{author}{Z.~Weng}, \bibinfo{author}{S.~Liu}, \bibinfo{author}{Z.~Zheng}, \bibinfo{author}{Y.~Zhang}, \bibinfo{author}{C.~Gong},
\newblock \bibinfo{title}{Cattle facial matching recognition algorithm based on multi-view feature fusion},
\newblock \bibinfo{journal}{Electronics 2023, Vol. 12, Page 156} \bibinfo{volume}{12} (\bibinfo{year}{2022}) \bibinfo{pages}{156}. \URLprefix \url{https://www.mdpi.com/2079-9292/12/1/156/htm https://www.mdpi.com/2079-9292/12/1/156}. \DOIprefix\doi{10.3390/ELECTRONICS12010156}.
\bibitem[{Xu et~al.(2022)Xu, Pan, and Gao}]{Xu2022-2}
\bibinfo{author}{F.~Xu}, \bibinfo{author}{X.~Pan}, \bibinfo{author}{J.~Gao},
\newblock \bibinfo{title}{Feature fusion capsule network for cow face recognition},
\newblock \bibinfo{journal}{https://doi.org/10.1117/1.JEI.31.6.061817} \bibinfo{volume}{31} (\bibinfo{year}{2022}) \bibinfo{pages}{061817}. \URLprefix \url{https://www.spiedigitallibrary.org/journals/journal-of-electronic-imaging/volume-31/issue-6/061817/Feature-fusion-capsule-network-for-cow-face-recognition/10.1117/1.JEI.31.6.061817.full https://www.spiedigitallibrary.org/journals/journal-of-electronic-imaging/volume-31/issue-6/061817/Feature-fusion-capsule-network-for-cow-face-recognition/10.1117/1.JEI.31.6.061817.short}. \DOIprefix\doi{10.1117/1.JEI.31.6.061817}.
\bibitem[{Yang et~al.(2023)Yang, Xu, Zhao, and Song}]{Yang2023}
\bibinfo{author}{L.~Yang}, \bibinfo{author}{X.~Xu}, \bibinfo{author}{J.~Zhao}, \bibinfo{author}{H.~Song},
\newblock \bibinfo{title}{Fusion of retinaface and improved facenet for individual cow identification in natural scenes},
\newblock \bibinfo{journal}{Information Processing in Agriculture}  (\bibinfo{year}{2023}). \URLprefix \url{https://linkinghub.elsevier.com/retrieve/pii/S2214317323000653}. \DOIprefix\doi{10.1016/J.INPA.2023.09.001}.
\bibitem[{Zhang et~al.(2009)Zhang, Zhao, Kong, Zhang, Zhao, and Kong}]{Zhang2009}
\bibinfo{author}{M.~Zhang}, \bibinfo{author}{L.~Zhao}, \bibinfo{author}{Q.~Kong}, \bibinfo{author}{M.~Zhang}, \bibinfo{author}{L.~Zhao}, \bibinfo{author}{Q.~Kong},
\newblock \bibinfo{title}{An iris localization algorithm based on geometrical features of cow eyes},
\newblock \bibinfo{journal}{SPIE} \bibinfo{volume}{7495} (\bibinfo{year}{2009}) \bibinfo{pages}{749517}. \URLprefix \url{https://ui.adsabs.harvard.edu/abs/2009SPIE.7495E..17Z/abstract}. \DOIprefix\doi{10.1117/12.832494}.
\bibitem[{Zhao et~al.(2011)Zhao, Shengnan, and Wang}]{Zhao2011}
\bibinfo{author}{L.~Zhao}, \bibinfo{author}{S.~Shengnan}, \bibinfo{author}{X.~Wang},
\newblock \bibinfo{title}{Tracking and traceability system using livestock iris code in meat supply chain},
\newblock \bibinfo{journal}{International Journal of Innovative Computing, Information and Control} \bibinfo{volume}{7} (\bibinfo{year}{2011}) \bibinfo{pages}{2201--2212}.
\bibitem[{Larregui et~al.(2019)Larregui, Cazzato, and Castro}]{Larregui2019}
\bibinfo{author}{J.~I. Larregui}, \bibinfo{author}{D.~Cazzato}, \bibinfo{author}{S.~M. Castro},
\newblock \bibinfo{title}{An image processing pipeline to segment iris for unconstrained cow identification system},
\newblock \bibinfo{journal}{Open Computer Science} \bibinfo{volume}{9} (\bibinfo{year}{2019}) \bibinfo{pages}{145--159}. \URLprefix \url{https://www.degruyter.com/document/doi/10.1515/comp-2019-0010/html?lang=en}. \DOIprefix\doi{10.1515/COMP-2019-0010/MACHINEREADABLECITATION/RIS}.
\bibitem[{Larregui and Cazzato(2019)}]{LarreguiD2019}
\bibinfo{author}{S.~M. C. J.~I. Larregui}, \bibinfo{author}{D.~Cazzato}, \bibinfo{title}{juanilarregui/bovineaaeyes80: Bovineaaeyes80 dataset for bovine biometric iris segmentation}, \bibinfo{year}{2019}. \URLprefix \url{https://github.com/juanilarregui/BovineAAEyes80}. \DOIprefix\doi{10.1016/J.COMPAG.2019.104944}.
\bibitem[{Li et~al.(2022)Li, Erickson, and Xiong}]{Li2022}
\bibinfo{author}{G.~Li}, \bibinfo{author}{G.~E. Erickson}, \bibinfo{author}{Y.~Xiong},
\newblock \bibinfo{title}{Individual beef cattle identification using muzzle images and deep learning techniques},
\newblock \bibinfo{journal}{Animals 2022, Vol. 12, Page 1453} \bibinfo{volume}{12} (\bibinfo{year}{2022}) \bibinfo{pages}{1453}. \URLprefix \url{https://www.mdpi.com/2076-2615/12/11/1453/htm https://www.mdpi.com/2076-2615/12/11/1453}. \DOIprefix\doi{10.3390/ANI12111453}.
\bibitem[{Xiong et~al.(2022)Xiong, Li, and Erickson}]{xiong_2022_6324361}
\bibinfo{author}{Y.~Xiong}, \bibinfo{author}{G.~Li}, \bibinfo{author}{G.~Erickson}, \bibinfo{title}{{Beef Cattle Muzzle/Noseprint database for individual identification}}, \bibinfo{year}{2022}. \URLprefix \url{https://doi.org/10.5281/zenodo.6324361}. \DOIprefix\doi{10.5281/zenodo.6324361}.
\bibitem[{Cihan et~al.(2023)Cihan, Saygili, Ozmen, and Akyuzlu}]{Cihan2023}
\bibinfo{author}{P.~Cihan}, \bibinfo{author}{A.~Saygili}, \bibinfo{author}{N.~E. Ozmen}, \bibinfo{author}{M.~Akyuzlu},
\newblock \bibinfo{title}{Identification and recognition of animals from biometric markers using computer vision approaches: A review},
\newblock \bibinfo{journal}{Kafkas Universitesi Veteriner Fakultesi Dergisi} \bibinfo{volume}{29} (\bibinfo{year}{2023}) \bibinfo{pages}{581--593}. \DOIprefix\doi{10.9775/KVFD.2023.30265}.
\bibitem[{Hao et~al.(2023)Hao, Zhang, Han, Hao, Wang, Li, and Liu}]{Hao2023}
\bibinfo{author}{W.~Hao}, \bibinfo{author}{K.~Zhang}, \bibinfo{author}{M.~Han}, \bibinfo{author}{W.~Hao}, \bibinfo{author}{J.~Wang}, \bibinfo{author}{F.~Li}, \bibinfo{author}{Z.~Liu},
\newblock \bibinfo{title}{A novel jinnan individual cattle recognition approach based on mutual attention learning scheme},
\newblock \bibinfo{journal}{Expert Systems with Applications} \bibinfo{volume}{230} (\bibinfo{year}{2023}) \bibinfo{pages}{120551}. \DOIprefix\doi{10.1016/J.ESWA.2023.120551}.
\bibitem[{Qiao et~al.(2021)Qiao, Kong, Clark, Lomax, Su, Eiffert, and Sukkarieh}]{Qiao2021}
\bibinfo{author}{Y.~Qiao}, \bibinfo{author}{H.~Kong}, \bibinfo{author}{C.~Clark}, \bibinfo{author}{S.~Lomax}, \bibinfo{author}{D.~Su}, \bibinfo{author}{S.~Eiffert}, \bibinfo{author}{S.~Sukkarieh},
\newblock \bibinfo{title}{Intelligent perception for cattle monitoring: A review for cattle identification, body condition score evaluation, and weight estimation},
\newblock \bibinfo{journal}{Computers and Electronics in Agriculture} \bibinfo{volume}{185} (\bibinfo{year}{2021}) \bibinfo{pages}{106143}. \DOIprefix\doi{10.1016/J.COMPAG.2021.106143}.
\bibitem[{Gao et~al.(2022)Gao, Burghardt, and Campbell}]{Gao2022}
\bibinfo{author}{J.~Gao}, \bibinfo{author}{T.~Burghardt}, \bibinfo{author}{N.~W. Campbell},
\newblock \bibinfo{title}{Label a herd in minutes: Individual holstein-friesian cattle identification},
\newblock \bibinfo{journal}{Lecture Notes in Computer Science (including subseries Lecture Notes in Artificial Intelligence and Lecture Notes in Bioinformatics)} \bibinfo{volume}{13374 LNCS} (\bibinfo{year}{2022}) \bibinfo{pages}{384--396}. \URLprefix \url{https://link.springer.com/chapter/10.1007/978-3-031-13324-4_33}. \DOIprefix\doi{10.1007/978-3-031-13324-4_33/FIGURES/7}.
\bibitem[{Campbell et~al.(2021)Campbell, Burghardt, Gao, Andrew, and Dowsey}]{Neill2021}
\bibinfo{author}{N.~Campbell}, \bibinfo{author}{T.~Burghardt}, \bibinfo{author}{J.~Gao}, \bibinfo{author}{W.~Andrew}, \bibinfo{author}{A.~Dowsey},
\newblock \bibinfo{title}{Cows2021 dataset},
\newblock \bibinfo{journal}{arXiv preprint arXiv:2105.01938}  (\bibinfo{year}{2021}). \DOIprefix\doi{10.5523/BRIS.4VNRCA7QW1642QLWXJADP87H7}.
\bibitem[{Gao et~al.(2021)Gao, Burghardt, Andrew, Dowsey, and Campbell}]{Gao2021}
\bibinfo{author}{J.~Gao}, \bibinfo{author}{T.~Burghardt}, \bibinfo{author}{W.~Andrew}, \bibinfo{author}{A.~W. Dowsey}, \bibinfo{author}{N.~W. Campbell},
\newblock \bibinfo{title}{Towards self-supervision for video identification of individual holstein-friesian cattle: The cows2021 dataset},
\newblock \bibinfo{journal}{Event Conference on Computer Vision and Pattern Recognition Workshop on Computer Vision for Animal Behavior Tracking and Modeling (CV4Animals)}  (\bibinfo{year}{2021}). \URLprefix \url{https://research-information.bris.ac.uk/en/publications/towards-self-supervision-for-video-identification-of-individual-h}.
\bibitem[{Andrew et~al.(2020)Andrew, Burghardt, Campbell, and Gao}]{OpenCows2020}
\bibinfo{author}{W.~Andrew}, \bibinfo{author}{T.~Burghardt}, \bibinfo{author}{N.~Campbell}, \bibinfo{author}{J.~Gao},
\newblock \bibinfo{title}{Opencows2020},
\newblock \bibinfo{journal}{Computers and Electronics in Agriculture}  (\bibinfo{year}{2020}). \DOIprefix\doi{10.5523/BRIS.10M32XL88X2B61ZLKKGZ3FML17}.
\bibitem[{Andrew et~al.(2017{\natexlab{a}})Andrew, Greatwood, and Burghardt}]{Andrew2017}
\bibinfo{author}{W.~Andrew}, \bibinfo{author}{C.~Greatwood}, \bibinfo{author}{T.~Burghardt},
\newblock \bibinfo{title}{Visual localisation and individual identification of holstein friesian cattle via deep learning},
\newblock \bibinfo{journal}{Proceedings - 2017 IEEE International Conference on Computer Vision Workshops, ICCVW 2017} \bibinfo{volume}{2018-January} (\bibinfo{year}{2017}{\natexlab{a}}) \bibinfo{pages}{2850--2859}. \DOIprefix\doi{10.1109/ICCVW.2017.336}.
\bibitem[{Andrew et~al.(2017{\natexlab{b}})Andrew, Greatwood, and Burghardt}]{AerialCattle2017}
\bibinfo{author}{W.~Andrew}, \bibinfo{author}{C.~Greatwood}, \bibinfo{author}{T.~Burghardt},
\newblock \bibinfo{title}{Aerialcattle2017},
\newblock \bibinfo{journal}{Proceedings - 2017 IEEE International Conference on Computer Vision Workshops, ICCVW 2017}  (\bibinfo{year}{2017}{\natexlab{b}}). \DOIprefix\doi{10.5523/BRIS.3OWFLKU95BXSX24643CYBXU3QH}.
\bibitem[{Andrew et~al.(2019)Andrew, Greatwood, and Burghardt}]{Andrew2019}
\bibinfo{author}{W.~Andrew}, \bibinfo{author}{C.~Greatwood}, \bibinfo{author}{T.~Burghardt},
\newblock \bibinfo{title}{Aerial animal biometrics: Individual friesian cattle recovery and visual identification via an autonomous uav with onboard deep inference},
\newblock \bibinfo{journal}{IEEE International Conference on Intelligent Robots and Systems}  (\bibinfo{year}{2019}) \bibinfo{pages}{237--243}. \DOIprefix\doi{10.1109/IROS40897.2019.8968555}.
\bibitem[{Zhao and Lian(2022)}]{Zhao2022}
\bibinfo{author}{J.~M. Zhao}, \bibinfo{author}{Q.~S. Lian},
\newblock \bibinfo{title}{Compact loss for visual identification of cattle in the wild},
\newblock \bibinfo{journal}{Computers and Electronics in Agriculture} \bibinfo{volume}{195} (\bibinfo{year}{2022}) \bibinfo{pages}{106784}. \DOIprefix\doi{10.1016/J.COMPAG.2022.106784}.
\bibitem[{Lu et~al.(2023)Lu, Weng, Zheng, Zhang, and Gong}]{Lu2023}
\bibinfo{author}{Y.~Lu}, \bibinfo{author}{Z.~Weng}, \bibinfo{author}{Z.~Zheng}, \bibinfo{author}{Y.~Zhang}, \bibinfo{author}{C.~Gong},
\newblock \bibinfo{title}{Algorithm for cattle identification based on locating key area},
\newblock \bibinfo{journal}{Expert Systems with Applications} \bibinfo{volume}{228} (\bibinfo{year}{2023}) \bibinfo{pages}{120365}. \DOIprefix\doi{10.1016/J.ESWA.2023.120365}.
\bibitem[{Hansen et~al.(2018)Hansen, Smith, Smith, Jabbar, and Forbes}]{Hansen2018}
\bibinfo{author}{M.~F. Hansen}, \bibinfo{author}{M.~L. Smith}, \bibinfo{author}{L.~N. Smith}, \bibinfo{author}{K.~A. Jabbar}, \bibinfo{author}{D.~Forbes},
\newblock \bibinfo{title}{Automated monitoring of dairy cow body condition, mobility and weight using a single 3d video capture device},
\newblock \bibinfo{journal}{Computers in industry} \bibinfo{volume}{98} (\bibinfo{year}{2018}) \bibinfo{pages}{14--22}.
\bibitem[{Alvarez et~al.(2018)Alvarez, Arroqui, Mangudo, Toloza, Jatip, Rodr{\'\i}guez, Teyseyre, Sanz, Zunino, Machado et~al.}]{Alvarez2018}
\bibinfo{author}{J.~R. Alvarez}, \bibinfo{author}{M.~Arroqui}, \bibinfo{author}{P.~Mangudo}, \bibinfo{author}{J.~Toloza}, \bibinfo{author}{D.~Jatip}, \bibinfo{author}{J.~M. Rodr{\'\i}guez}, \bibinfo{author}{A.~Teyseyre}, \bibinfo{author}{C.~Sanz}, \bibinfo{author}{A.~Zunino}, \bibinfo{author}{C.~Machado}, et~al.,
\newblock \bibinfo{title}{Body condition estimation on cows from depth images using convolutional neural networks},
\newblock \bibinfo{journal}{Computers and electronics in agriculture} \bibinfo{volume}{155} (\bibinfo{year}{2018}) \bibinfo{pages}{12--22}.
\bibitem[{Yukun et~al.(2019)Yukun, Pengju, Yujie, Ziqi, Yang, Baisheng, Runze, and Yonggen}]{Yukun2019}
\bibinfo{author}{S.~Yukun}, \bibinfo{author}{H.~Pengju}, \bibinfo{author}{W.~Yujie}, \bibinfo{author}{C.~Ziqi}, \bibinfo{author}{L.~Yang}, \bibinfo{author}{D.~Baisheng}, \bibinfo{author}{L.~Runze}, \bibinfo{author}{Z.~Yonggen},
\newblock \bibinfo{title}{Automatic monitoring system for individual dairy cows based on a deep learning framework that provides identification via body parts and estimation of body condition score},
\newblock \bibinfo{journal}{Journal of dairy science} \bibinfo{volume}{102} (\bibinfo{year}{2019}) \bibinfo{pages}{10140--10151}.
\bibitem[{Huang et~al.(2019)Huang, Li, and Hu}]{Huang2019}
\bibinfo{author}{X.~Huang}, \bibinfo{author}{X.~Li}, \bibinfo{author}{Z.~Hu},
\newblock \bibinfo{title}{Cow tail detection method for body condition score using faster r-cnn},
\newblock in: \bibinfo{booktitle}{2019 IEEE International Conference on Unmanned Systems and Artificial Intelligence (ICUSAI)}, \bibinfo{organization}{IEEE}, \bibinfo{year}{2019}, pp. \bibinfo{pages}{347--351}.
\bibitem[{Fischer et~al.(2015)Fischer, Luginb{\"u}hl, Delattre, Delouard, and Faverdin}]{Fischer2015}
\bibinfo{author}{A.~Fischer}, \bibinfo{author}{T.~Luginb{\"u}hl}, \bibinfo{author}{L.~Delattre}, \bibinfo{author}{J.~Delouard}, \bibinfo{author}{P.~Faverdin},
\newblock \bibinfo{title}{Rear shape in 3 dimensions summarized by principal component analysis is a good predictor of body condition score in holstein dairy cows},
\newblock \bibinfo{journal}{Journal of dairy science} \bibinfo{volume}{98} (\bibinfo{year}{2015}) \bibinfo{pages}{4465--4476}.
\bibitem[{Liu et~al.(2020)Liu, He, and Norton}]{Liu2020}
\bibinfo{author}{D.~Liu}, \bibinfo{author}{D.~He}, \bibinfo{author}{T.~Norton},
\newblock \bibinfo{title}{Automatic estimation of dairy cattle body condition score from depth image using ensemble model},
\newblock \bibinfo{journal}{biosystems engineering} \bibinfo{volume}{194} (\bibinfo{year}{2020}) \bibinfo{pages}{16--27}.
\bibitem[{Kadlec et~al.(2022)Kadlec, Indest, Castro, Waqar, Campos, Amorim, Bi, Hanigan, and Morota}]{Kadlec2022}
\bibinfo{author}{R.~Kadlec}, \bibinfo{author}{S.~Indest}, \bibinfo{author}{K.~Castro}, \bibinfo{author}{S.~Waqar}, \bibinfo{author}{L.~M. Campos}, \bibinfo{author}{S.~T. Amorim}, \bibinfo{author}{Y.~Bi}, \bibinfo{author}{M.~D. Hanigan}, \bibinfo{author}{G.~Morota},
\newblock \bibinfo{title}{Automated acquisition of top-view dairy cow depth image data using an rgb-d sensor camera},
\newblock \bibinfo{journal}{Translational Animal Science} \bibinfo{volume}{6} (\bibinfo{year}{2022}). \URLprefix \url{https://dx.doi.org/10.1093/tas/txac163}. \DOIprefix\doi{10.1093/TAS/TXAC163}.
\bibitem[{Han et~al.(2013)Han, Shao, Xu, and Shotton}]{Han2013}
\bibinfo{author}{J.~Han}, \bibinfo{author}{L.~Shao}, \bibinfo{author}{D.~Xu}, \bibinfo{author}{J.~Shotton},
\newblock \bibinfo{title}{Enhanced computer vision with microsoft kinect sensor: A review},
\newblock \bibinfo{journal}{IEEE Transactions on Cybernetics} \bibinfo{volume}{43} (\bibinfo{year}{2013}) \bibinfo{pages}{1318--1334}. \DOIprefix\doi{10.1109/TCYB.2013.2265378}.
\bibitem[{Andrew et~al.(2016)Andrew, Hannuna, Campbell, and Burghardt}]{Andrew2016}
\bibinfo{author}{W.~Andrew}, \bibinfo{author}{S.~Hannuna}, \bibinfo{author}{N.~Campbell}, \bibinfo{author}{T.~Burghardt},
\newblock \bibinfo{title}{Automatic individual holstein friesian cattle identification via selective local coat pattern matching in rgb-d imagery},
\newblock \bibinfo{journal}{Proceedings - International Conference on Image Processing, ICIP} \bibinfo{volume}{2016-August} (\bibinfo{year}{2016}) \bibinfo{pages}{484--488}. \DOIprefix\doi{10.1109/ICIP.2016.7532404}.
\bibitem[{Pedregosa et~al.(2011)Pedregosa, Varoquaux, Gramfort, Michel, Thirion, Grisel, Blondel, Prettenhofer, Weiss, Dubourg, Vanderplas, Passos, Cournapeau, Brucher, Perrot, and Duchesnay}]{scikit-learn}
\bibinfo{author}{F.~Pedregosa}, \bibinfo{author}{G.~Varoquaux}, \bibinfo{author}{A.~Gramfort}, \bibinfo{author}{V.~Michel}, \bibinfo{author}{B.~Thirion}, \bibinfo{author}{O.~Grisel}, \bibinfo{author}{M.~Blondel}, \bibinfo{author}{P.~Prettenhofer}, \bibinfo{author}{R.~Weiss}, \bibinfo{author}{V.~Dubourg}, \bibinfo{author}{J.~Vanderplas}, \bibinfo{author}{A.~Passos}, \bibinfo{author}{D.~Cournapeau}, \bibinfo{author}{M.~Brucher}, \bibinfo{author}{M.~Perrot}, \bibinfo{author}{E.~Duchesnay},
\newblock \bibinfo{title}{Scikit-learn: Machine learning in {P}ython},
\newblock \bibinfo{journal}{Journal of Machine Learning Research} \bibinfo{volume}{12} (\bibinfo{year}{2011}) \bibinfo{pages}{2825--2830}.
\bibitem[{Hossain and Lin(2023)}]{Hossain2023}
\bibinfo{author}{S.~Hossain}, \bibinfo{author}{X.~Lin},
\newblock \bibinfo{title}{Efficient stereo depth estimation for pseudo-lidar: A self-supervised approach based on multi-input resnet encoder},
\newblock \bibinfo{journal}{Sensors} \bibinfo{volume}{23} (\bibinfo{year}{2023}). \DOIprefix\doi{10.3390/S23031650}.
\bibitem[{Eldar et~al.(1994)Eldar, Lindenbaum, Porat, and Zeevi}]{Eldar1994}
\bibinfo{author}{Y.~Eldar}, \bibinfo{author}{M.~Lindenbaum}, \bibinfo{author}{M.~Porat}, \bibinfo{author}{Y.~Y. Zeevi},
\newblock \bibinfo{title}{The farthest point strategy for progressive image sampling},
\newblock \bibinfo{journal}{Proceedings - International Conference on Pattern Recognition} \bibinfo{volume}{3} (\bibinfo{year}{1994}) \bibinfo{pages}{93--97}. \DOIprefix\doi{10.1109/ICPR.1994.577129}.
\bibitem[{Qi et~al.(2016)Qi, Su, Mo, and Guibas}]{Qi2016}
\bibinfo{author}{C.~R. Qi}, \bibinfo{author}{H.~Su}, \bibinfo{author}{K.~Mo}, \bibinfo{author}{L.~J. Guibas},
\newblock \bibinfo{title}{Pointnet: Deep learning on point sets for 3d classification and segmentation},
\newblock \bibinfo{journal}{Proceedings - 30th IEEE Conference on Computer Vision and Pattern Recognition, CVPR 2017} \bibinfo{volume}{2017-January} (\bibinfo{year}{2016}) \bibinfo{pages}{77--85}. \URLprefix \url{https://arxiv.org/abs/1612.00593v2}. \DOIprefix\doi{10.1109/CVPR.2017.16}.
\bibitem[{Rusu et~al.(2009)Rusu, Blodow, and Beetz}]{Rusu2009}
\bibinfo{author}{R.~B. Rusu}, \bibinfo{author}{N.~Blodow}, \bibinfo{author}{M.~Beetz},
\newblock \bibinfo{title}{Fast point feature histograms (fpfh) for 3d registration},
\newblock \bibinfo{journal}{Proceedings - IEEE International Conference on Robotics and Automation}  (\bibinfo{year}{2009}) \bibinfo{pages}{3212--3217}. \DOIprefix\doi{10.1109/ROBOT.2009.5152473}.
\bibitem[{Rusu et~al.(2010)Rusu, Bradski, Thibaux, and Hsu}]{Rusu2010}
\bibinfo{author}{R.~B. Rusu}, \bibinfo{author}{G.~Bradski}, \bibinfo{author}{R.~Thibaux}, \bibinfo{author}{J.~Hsu},
\newblock \bibinfo{title}{Fast 3d recognition and pose using the viewpoint feature histogram},
\newblock \bibinfo{journal}{IEEE/RSJ 2010 International Conference on Intelligent Robots and Systems, IROS 2010 - Conference Proceedings}  (\bibinfo{year}{2010}) \bibinfo{pages}{2155--2162}. \DOIprefix\doi{10.1109/IROS.2010.5651280}.
\bibitem[{Marton et~al.(2010)Marton, Blodow, Beetz, Marton, Pangercic, and Kleinehellefort}]{Marton2010}
\bibinfo{author}{Z.~C. Marton}, \bibinfo{author}{N.~Blodow}, \bibinfo{author}{M.~Beetz}, \bibinfo{author}{Z.-C. Marton}, \bibinfo{author}{D.~Pangercic}, \bibinfo{author}{J.~Kleinehellefort},
\newblock \bibinfo{title}{General 3d modelling of novel objects from a single view},
\newblock \bibinfo{journal}{ieeexplore.ieee.orgZC Marton, D Pangercic, N Blodow, J Kleinehellefort, M Beetz2010 ieee/rsj international conference on intelligent robots and, 2010•ieeexplore.ieee.org}  (\bibinfo{year}{2010}). \URLprefix \url{https://ieeexplore.ieee.org/abstract/document/5650434/}. \DOIprefix\doi{10.1109/IROS.2010.5650434}.
\bibitem[{Wu et~al.(2017)Wu, Li, Ranasinghe, Dissanayake, and Liu}]{Wu2017}
\bibinfo{author}{K.~Wu}, \bibinfo{author}{X.~Li}, \bibinfo{author}{R.~Ranasinghe}, \bibinfo{author}{G.~Dissanayake}, \bibinfo{author}{Y.~Liu},
\newblock \bibinfo{title}{Risas: A novel rotation, illumination, scale invariant appearance and shape feature},
\newblock \bibinfo{journal}{Proceedings - IEEE International Conference on Robotics and Automation}  (\bibinfo{year}{2017}) \bibinfo{pages}{4008--4015}. \DOIprefix\doi{10.1109/ICRA.2017.7989461}.
\bibitem[{Han et~al.(2023)Han, Feng, Sun, and Xiao}]{Han2023}
\bibinfo{author}{X.~F. Han}, \bibinfo{author}{Z.~A. Feng}, \bibinfo{author}{S.~J. Sun}, \bibinfo{author}{G.~Q. Xiao},
\newblock \bibinfo{title}{3d point cloud descriptors: state-of-the-art},
\newblock \bibinfo{journal}{Artificial Intelligence Review} \bibinfo{volume}{56} (\bibinfo{year}{2023}) \bibinfo{pages}{12033--12083}. \URLprefix \url{https://link.springer.com/article/10.1007/s10462-023-10486-4}. \DOIprefix\doi{10.1007/S10462-023-10486-4/TABLES/6}.
\bibitem[{He et~al.(2016)He, Zhang, Ren, and Sun}]{He2016}
\bibinfo{author}{K.~He}, \bibinfo{author}{X.~Zhang}, \bibinfo{author}{S.~Ren}, \bibinfo{author}{J.~Sun},
\newblock \bibinfo{title}{Deep residual learning for image recognition},
\newblock \bibinfo{journal}{Proceedings of the IEEE Computer Society Conference on Computer Vision and Pattern Recognition} \bibinfo{volume}{2016-December} (\bibinfo{year}{2016}) \bibinfo{pages}{770--778}. \DOIprefix\doi{10.1109/CVPR.2016.90}.
\bibitem[{Yang et~al.(2019)Yang, Mirmehdi, and Burghardt}]{Yang2019}
\bibinfo{author}{X.~Yang}, \bibinfo{author}{M.~Mirmehdi}, \bibinfo{author}{T.~Burghardt},
\newblock \bibinfo{title}{Great ape detection in challenging jungle camera trap footage via attention-based spatial and temporal feature blending},
\newblock \bibinfo{journal}{2019 IEEE/CVF International Conference on Computer Vision Workshop (ICCVW)}  (\bibinfo{year}{2019}) \bibinfo{pages}{255--262}. \DOIprefix\doi{10.1109/ICCVW.2019.00034}.
\bibitem[{Hadsell et~al.(2006)Hadsell, Chopra, and LeCun}]{Hadsell2006}
\bibinfo{author}{R.~Hadsell}, \bibinfo{author}{S.~Chopra}, \bibinfo{author}{Y.~LeCun},
\newblock \bibinfo{title}{Dimensionality reduction by learning an invariant mapping},
\newblock \bibinfo{journal}{Proceedings of the IEEE Computer Society Conference on Computer Vision and Pattern Recognition} \bibinfo{volume}{2} (\bibinfo{year}{2006}) \bibinfo{pages}{1735--1742}. \DOIprefix\doi{10.1109/CVPR.2006.100}.
\bibitem[{Schroff et~al.(2015)Schroff, Kalenichenko, and Philbin}]{Schroff2015}
\bibinfo{author}{F.~Schroff}, \bibinfo{author}{D.~Kalenichenko}, \bibinfo{author}{J.~Philbin},
\newblock \bibinfo{title}{Facenet: A unified embedding for face recognition and clustering},
\newblock \bibinfo{journal}{Proceedings of the IEEE Computer Society Conference on Computer Vision and Pattern Recognition} \bibinfo{volume}{07-12-June-2015} (\bibinfo{year}{2015}) \bibinfo{pages}{815--823}. \URLprefix \url{https://arxiv.org/abs/1503.03832v3}. \DOIprefix\doi{10.1109/cvpr.2015.7298682}.
\bibitem[{Masullo et~al.(2019)Masullo, Burghardt, Damen, Perrett, and Mirmehdi}]{Masullo2019}
\bibinfo{author}{A.~Masullo}, \bibinfo{author}{T.~Burghardt}, \bibinfo{author}{D.~Damen}, \bibinfo{author}{T.~Perrett}, \bibinfo{author}{M.~Mirmehdi},
\newblock \bibinfo{title}{Who goes there? exploiting silhouettes and wearable signals for subject identification in multi-person environments},
\newblock \bibinfo{journal}{Proceedings - 2019 International Conference on Computer Vision Workshop, ICCVW 2019}  (\bibinfo{year}{2019}) \bibinfo{pages}{1599--1607}. \DOIprefix\doi{10.1109/ICCVW.2019.00199}.
\bibitem[{Wang et~al.(2017)Wang, Eliott, Ainooson, Palmer, and Kunda}]{Wang2017}
\bibinfo{author}{X.~Wang}, \bibinfo{author}{F.~M. Eliott}, \bibinfo{author}{J.~Ainooson}, \bibinfo{author}{J.~H. Palmer}, \bibinfo{author}{M.~Kunda},
\newblock \bibinfo{title}{An object is worth six thousand pictures: The egocentric, manual, multi-image (emmi) dataset},
\newblock \bibinfo{journal}{Proceedings - 2017 IEEE International Conference on Computer Vision Workshops, ICCVW 2017} \bibinfo{volume}{2018-January} (\bibinfo{year}{2017}) \bibinfo{pages}{2364--2372}. \DOIprefix\doi{10.1109/ICCVW.2017.279}.
\bibitem[{Shorten and Khoshgoftaar(2019)}]{Shorten2019}
\bibinfo{author}{C.~Shorten}, \bibinfo{author}{T.~M. Khoshgoftaar},
\newblock \bibinfo{title}{A survey on image data augmentation for deep learning},
\newblock \bibinfo{journal}{Journal of Big Data} \bibinfo{volume}{6} (\bibinfo{year}{2019}) \bibinfo{pages}{1--48}. \URLprefix \url{https://journalofbigdata.springeropen.com/articles/10.1186/s40537-019-0197-0}. \DOIprefix\doi{10.1186/S40537-019-0197-0/FIGURES/33}.
\bibitem[{Bloice(2022)}]{Bloice2022}
\bibinfo{author}{M.~D. Bloice}, \bibinfo{title}{mdbloice/augmentor: Image augmentation library in python for machine learning.}, \bibinfo{year}{2022}. \URLprefix \url{https://github.com/mdbloice/Augmentor}.
\bibitem[{Buslaev et~al.(2020)Buslaev, Iglovikov, Khvedchenya, Parinov, Druzhinin, and Kalinin}]{Buslaev2020}
\bibinfo{author}{A.~Buslaev}, \bibinfo{author}{V.~I. Iglovikov}, \bibinfo{author}{E.~Khvedchenya}, \bibinfo{author}{A.~Parinov}, \bibinfo{author}{M.~Druzhinin}, \bibinfo{author}{A.~A. Kalinin},
\newblock \bibinfo{title}{Albumentations: Fast and flexible image augmentations},
\newblock \bibinfo{journal}{Information} \bibinfo{volume}{11} (\bibinfo{year}{2020}). \URLprefix \url{https://www.mdpi.com/2078-2489/11/2/125}. \DOIprefix\doi{10.3390/info11020125}.
\bibitem[{Advanced Computing Research Centre, University of Bristol(2023)}]{bc4}
Advanced Computing Research Centre, University of Bristol, \bibinfo{title}{Advanced computing research centre}, \bibinfo{year}{2023}. \URLprefix \url{https://www.acrc.bris.ac.uk/acrc/phase4.htm}.
\bibitem[{Chen et~al.(2020)Chen, Chen, Hajimirsadeghi, and Mori}]{chen2020adapting}
\bibinfo{author}{L.~Chen}, \bibinfo{author}{J.~Chen}, \bibinfo{author}{H.~Hajimirsadeghi}, \bibinfo{author}{G.~Mori},
\newblock \bibinfo{title}{Adapting grad-cam for embedding networks},
\newblock in: \bibinfo{booktitle}{Proceedings of the IEEE/CVF Winter Conference on Applications of Computer Vision}, \bibinfo{year}{2020}, pp. \bibinfo{pages}{2794--2803}.
\bibitem[{Selvaraju et~al.(2017)Selvaraju, Cogswell, Das, Vedantam, Parikh, and Batra}]{selvaraju2017grad}
\bibinfo{author}{R.~R. Selvaraju}, \bibinfo{author}{M.~Cogswell}, \bibinfo{author}{A.~Das}, \bibinfo{author}{R.~Vedantam}, \bibinfo{author}{D.~Parikh}, \bibinfo{author}{D.~Batra},
\newblock \bibinfo{title}{Grad-cam: Visual explanations from deep networks via gradient-based localization},
\newblock in: \bibinfo{booktitle}{Proceedings of the IEEE international conference on computer vision}, \bibinfo{year}{2017}, pp. \bibinfo{pages}{618--626}.
\bibitem[{Zheng et~al.(2019)Zheng, Chen, Yuan, Li, and Ren}]{Zheng2019}
\bibinfo{author}{T.~Zheng}, \bibinfo{author}{C.~Chen}, \bibinfo{author}{J.~Yuan}, \bibinfo{author}{B.~Li}, \bibinfo{author}{K.~Ren},
\newblock \bibinfo{title}{Pointcloud saliency maps},
\newblock \bibinfo{journal}{2019 IEEE/CVF International Conference on Computer Vision (ICCV)} \bibinfo{volume}{2019-October} (\bibinfo{year}{2019}) \bibinfo{pages}{1598--1606}. \DOIprefix\doi{10.1109/ICCV.2019.00168}.
\bibitem[{Haider and Hel-Or(2022)}]{Haider2022}
\bibinfo{author}{A.~Haider}, \bibinfo{author}{H.~Hel-Or},
\newblock \bibinfo{title}{What can we learn from depth camera sensor noise?},
\newblock \bibinfo{journal}{Sensors} \bibinfo{volume}{22} (\bibinfo{year}{2022}) \bibinfo{pages}{5448}. \DOIprefix\doi{10.3390/s22145448}.
\bibitem[{McManus et~al.(2022)McManus, Boden, Weir, Viora, Barker, Kim, McBride, and Yang}]{McManus2022}
\bibinfo{author}{R.~McManus}, \bibinfo{author}{L.~A. Boden}, \bibinfo{author}{W.~Weir}, \bibinfo{author}{L.~Viora}, \bibinfo{author}{R.~Barker}, \bibinfo{author}{Y.~Kim}, \bibinfo{author}{P.~McBride}, \bibinfo{author}{S.~Yang},
\newblock \bibinfo{title}{Thermography for disease detection in livestock: A scoping review},
\newblock \bibinfo{journal}{Frontiers in Veterinary Science} \bibinfo{volume}{9} (\bibinfo{year}{2022}) \bibinfo{pages}{965622}. \DOIprefix\doi{10.3389/FVETS.2022.965622/BIBTEX}.

\end{thebibliography}
\printglossaries
\end{document}